%% file: iclr2025_conference.tex
\documentclass{article} 
\usepackage{iclr2025_conference,times}

\input{math_commands.tex}

\iclrfinalcopy
\usepackage{url}
\usepackage[utf8]{inputenc} 
\usepackage[T1]{fontenc}    
\usepackage[hidelinks]{hyperref}       
\usepackage{url}            
\usepackage{booktabs}       
\usepackage{amsfonts}       
\usepackage{amsmath}
\usepackage{amssymb}
\usepackage{nicefrac}       
\usepackage{microtype}      
\usepackage{xcolor}         
\usepackage{multirow}
\usepackage{caption}
\usepackage{subcaption}
\usepackage{multicol}
\usepackage{graphicx}
\usepackage{booktabs}
\usepackage[export]{adjustbox}
\usepackage{mathdots}

\usepackage[mode=buildmissing]{standalone}
\usepackage{tikz}
\usepackage{pgfplots}
\usepgfplotslibrary{groupplots,dateplot}
\usetikzlibrary{patterns,shapes.arrows,snakes}
\pgfplotsset{compat=newest}

\usepackage{comment}

\definecolor{tabBlue}{HTML}{1f77b4}
\definecolor{tabOrange}{HTML}{ff7f0e}
\definecolor{tabGreen}{HTML}{2ca02c}
\definecolor{tabRed}{HTML}{d62728}
\definecolor{tabPurple}{HTML}{9467bd}
\definecolor{tabBrown}{HTML}{8c564b}
\definecolor{tabPink}{HTML}{e377c2}
\definecolor{tabGray}{HTML}{7f7f7f}
\definecolor{tabOlive}{HTML}{bcbd22}
\definecolor{tabCyan}{HTML}{17becf}
\definecolor{blue}{HTML}{284ce0}

\DeclareUrlCommand\url{\color{magenta}}
\usepackage{acro}
\acsetup{make-links=true} 
\DeclareAcronym{gan}{
  short = GAN,
  long = Generative Adversarial neural Network,
}
\DeclareAcronym{cnn}{
    short = CNN,
    long = Convolutional Neural Network,
}
\DeclareAcronym{fwt}{
  short = FWT,
  long = Fast Wavelet Transform,
}
\DeclareAcronym{ifwt}{
  short = iFWT,
  long = inverse fast wavelet transform,
}
\DeclareAcronym{fft}{
    short = FFT,
    long = Fast Fourier Transform,
}
\DeclareAcronym{dct}{
  short = DCT,
  long = Discrete Cosine Transform,
}
\DeclareAcronym{ffhq}{
  short = FFHQ,
  long = Flickr Faces High Quality,
}
\DeclareAcronym{lsun}{
  short = LSUN,
  long = Large-scale Scene UNderstanding
}
\DeclareAcronym{celeba}{
  short = CelebA,
  long = Large-scale Celeb Faces Attributes
}
\DeclareAcronym{wpt}{
  short = $\mathcal{W}_p$,
  long = Wavelet Packet Transform
}
\DeclareAcronym{mse}{
  short = MSE,
  long = Mean Squared Error
}
\DeclareAcronym{fid}{
  short = FID,
  long = Fr\'echet Inception Distance
}
\DeclareAcronym{fd}{
  short = FD,
  long = Fr\'echet Distance
}
\DeclareAcronym{wpskl}{
  short = WPKL,
  long = Wavelet packet Power Kullback–Leibler Divergence
}
\DeclareAcronym{ddpm}{
  short = DDPM,
  long = Denoising Diffusion Probabilistic Models
}
\DeclareAcronym{ddim}{
  short = DDIM,
  long = Denoising Diffusion Implicit Models
}
\DeclareAcronym{wsgm}{
  short = WSGM,
  long = Wavelet Score Based Generative Model
}
\DeclareAcronym{is}{
  short = IS,
  long = Inception Score
}
\DeclareAcronym{mocap}{
  short = mocap,
  long = Motion capture
}
\DeclareAcronym{ddgan}{
  short = DDGAN,
  long = Denoising Diffusion GAN
}
\DeclareAcronym{ssim}{
  short = SSIM,
  long = Structural Similarity Index Measure
}
\DeclareAcronym{vae}{
  short = VAE,
  long = Variational AutoEncoder
}
\DeclareAcronym{dit}{
  short = DiT,
  long = Diffusion Transformer
}
\DeclareAcronym{kid}{
  short = KID,
  long = Kernel Inception Distance
}
\DeclareAcronym{fwd}{
  short = FWD,
  long = Fr\'echet Wavelet Distance
}
\DeclareAcronym{wavediff}{
  short = WaveDiff,
  long = Wavelet Diffusion
}
\DeclareAcronym{projgan}{
  short = Proj. FastGAN, 
  long = Projected Fast GAN,
}
\DeclareAcronym{celebahq}{
  short = CelebA-HQ,
  long = Large-scale Celeb Faces Attributes High Quality
}
\DeclareAcronym{agriculture}{
  short = DNDD-Dataset,
  long=Deep Nutrient Deficiency Dikopshof Dataset 
}
\DeclareAcronym{psnr}{
  short = PSNR,
  long = Peak Signal to Noise Ratio
}
\DeclareAcronym{cifar10}{
  short = CIFAR-10,
  long = Canadian Institute For Advanced Research-10
}
\DeclareAcronym{fdist}{
    short = FD,
    long = Fr\'echet Distance
}
\DeclareAcronym{fd_dino}{
    short = FD-DINOv2,
    long = DINOv2-Fr\'echet Distance
}
\DeclareAcronym{her}{
    short = HER,
    long = Human Error Rate 
}
\DeclareAcronym{swd}{
    short = SWD,
    long = Sliced Wasserstein Distance
}

\title{Fr\'echet Wavelet Distance: A Domain-Agnostic Metric for Image Generation}

\author{
Lokesh Veeramacheneni \\
University of Bonn \\
\textit{lveerama@uni-bonn.de}
\And
Moritz Wolter \\
University of Bonn \\
\textit{moritz.wolter@uni-bonn.de}
\And
Hildegard Kuehne \\
University of Tuebingen,\\MIT-IBM Watson AI Lab\\ 
\textit{h.kuehne@uni-tuebingen.de}\vspace{-5mm}
\And
Juergen Gall \\
University of Bonn, \\
Lamarr Institute for Machine Learning and Artificial Intelligence\\
\textit{gall@iai.uni-bonn.de}
}

\begin{document}

\maketitle

\begin{abstract}
   Modern metrics for generative learning like \ac{fid} and \ac*{fd_dino} demonstrate impressive performance. However, they suffer from various shortcomings, like a bias towards specific generators and datasets.
   To address this problem, we propose the \ac{fwd} as a domain-agnostic metric based on the \acf{wpt}.
   \ac{fwd} provides a sight across a broad spectrum of frequencies in images with a high resolution, preserving both spatial and textural aspects.
   Specifically, we use \ac{wpt} to project generated and real images to the packet coefficient space.
   We then compute the Fr\'echet distance with the resultant coefficients to evaluate the quality of a generator.
   This metric is general-purpose and dataset-domain agnostic, as it does not rely on any pre-trained network, while being more interpretable due to its ability to compute Fr\'echet distance per packet, enhancing transparency.
   We conclude with an extensive evaluation of a wide variety of generators across various datasets that the proposed \ac{fwd} can generalize and improve robustness to domain shifts and various corruptions compared to other metrics.
\end{abstract}

\section{Introduction}
\label{intro}
With the surge of generative neural networks, especially in the image domain, it becomes important to assess their performance in a robust and reliable way~\citep{heusel2017gans,binkowski18kid,salimans2016improved,Kynkaanniemi19improvpresrecall,stein2023exposing}.
\ac{fid}~\citep{heusel2017gans} has emerged as the de facto standard for comparing generative image synthesis approaches.
However, it also shows various shortcomings, such as its reliance on a pre-trained classification backbone, i.e., InceptionV3 trained on ImageNet.
This, by design, introduces a class dependency into \ac{fid} leading to accidental distortions \citep{Sauer2021NEURIPS}. The \ac{fid} scores improve if the evaluation set resembles ImageNet or if the use of an ImageNet pre-trained discriminator pushes the output distribution towards ImageNet, although the image quality remains the same in these cases \citep{Kynkaat2023TheRole}. 


\begin{figure}
  \centering
  \includegraphics[scale=0.3]{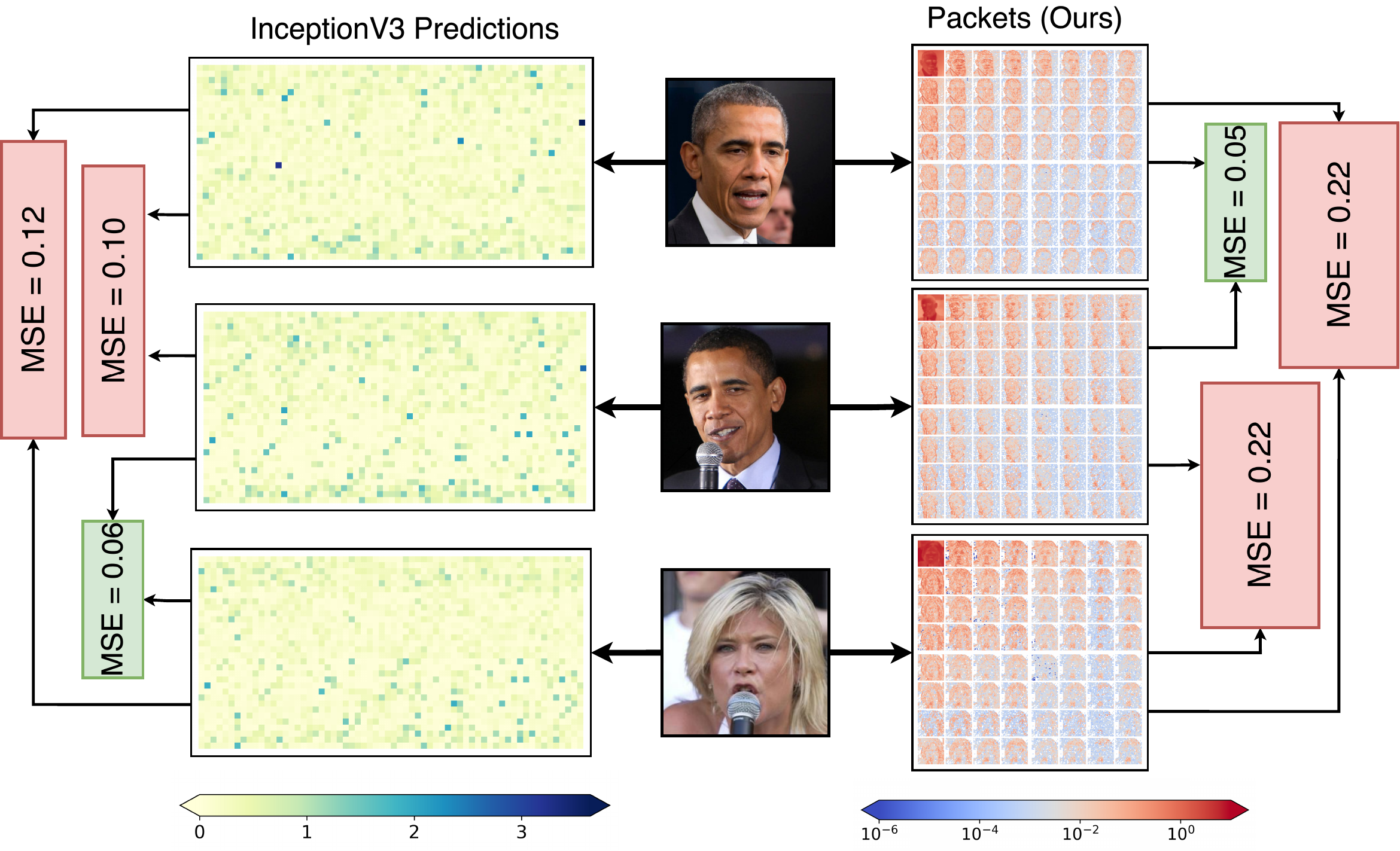}
  \caption{
  The first two images depict the same person, while the last image depicts a different person. Intuitively, the first two images are more similar than the other pairs of images. When computing the mean squared error between the images using the penultimate InceptionV3 activations or wavelet packets, we observe that the wavelet packets produce a low distance for the first two images, as expected. Surprisingly, according to InceptionV3, the last two images are similar since both images are classified as \texttt{`microphone'} whereas the first image as \texttt{`groom'}. Images from Flickr.}
  \label{fig:fid_bias}
\end{figure}

To address the domain bias problem caused by the use of a pre-trained network, we propose an alternative metric based on the \acf{wpt}. In contrast to other pure frequency \citep{manish12ftquality} or spatial \citep{ssim04wang,psnrvsssim2010alain} metrics, wavelets have the advantage that they combine both frequency and spatial aspects in one metric.
While frequency information is important \citep{durall2020watch,dzanic2020fourier,rahaman2019spectral,schwarz2021frequency,wolter2022wavelet}, it alone is insufficient to assess the quality of synthesized images without considering additional spatial information.
Wavelets are thus an ideal representation for a metric comparing generative approaches for image synthesis. 
As \ac{fid}, \ac{fwd} utilizes the Fr\'echet distance of the real and generated set of images as a distance measure, but it is not computed based on InceptionV3 activation maps. 
Instead, it utilizes the wavelet-packet frequency band representations of \ac{wpt} as illustrated in Figures~\ref{fig:fid_bias} and \ref{fig:fwd_formulation}. 
To this end, we first use \ac{wpt} to transform every image, where we use the Haar wavelet transform at a fixed level. We then compute the Fr\'echet distance for each packet of the transform and average them over all packets. The proposed \acf{fwd} thus considers spatial information as well as all frequency bands. 

To quantitatively assess those characteristics, we evaluate the proposed metric in terms of its domain bias and robustness.
We further compare the proposed \acs{fwd} to existing state-of-the-art metrics like \ac{fid}, \ac{kid}, and \ac{fd_dino} on standard datasets.
We show that \ac{fwd} is a more robust metric that does not suffer from the domain bias and can thus be applied to any dataset.
\cite{Kynkaat2023TheRole} experimented with optimizing \ac{fid} by selecting a subset of images from 250k generated images, where the subset's InceptionV3 activations are related to ImageNet classes. Building on this work, we observed a significant improvement in \ac{fid} by $\approx50\%$, when evaluated on this subset. \ac{fd_dino} responded to ImageNet-feature optimization with an improvement of $\approx2\%$ as well. This undesired improvement can likely be explained by the overlap between the ImageNet and the DINOv2 training set. In contrast, \ac{fwd} remains the same despite the manipulation. We also show that some unexpected \ac{fid} results can be attributed to the dataset bias.
Furthermore, \ac{fwd} is significantly faster to compute. The source code for computing \ac*{fwd} is available at:   \url{https://github.com/BonnBytes/PyTorch-FWD}.

In summary, this paper makes the following contributions: 
\begin{enumerate}
  \item We propose the \acf{fwd} as a dataset- and domain-agnostic metric for evaluation of generative approaches for image synthesis.
  \item \ac{fwd} is an interpretable metric, as the \acf{wpt} splits the frequency space into hierarchically organized, discrete subbands.
  \item We show that the proposed method is computationally inexpensive and robust to corruption, perturbation, and distractors.
  \item We show that \ac{fd_dino} addresses the domain bias issue to an extent but at a very high computational cost. Furthermore, we provide evidence that it is still limited to its training data domain.
\end{enumerate}
\section{Related Work}\label{sec:related_work}
\subsection{Metrics for generative learning}\label{sec:generative_metrics}

A generative model should generate novel image samples that mirror the training set sample distribution, including data diversity. In a vision context, \cite{salimans2016improved} proposed the \acf{is} as a measure of image quality, independent of the target dataset statistics. The \ac{is} is computed by measuring the entropy of the class probabilities of an InceptionV3.
The score builds upon the assumption that a generative network that has converged to a meaningful solution will produce images that will allow InceptionV3 to make predictions with certainty. In other words, a certain InceptionV3 has a low prediction entropy. \ac{is} has been found to be sensitive to different ImageNet training runs \citep{Barrat2018ISNote}. 
Furthermore, it does not use the statistics of the real data distribution a \ac{gan} is trained to model \citep{heusel2017gans}.
\cite{heusel2017gans} proposed \ac{fid} in response. Instead of measuring the entropy at the final layer, \ac{fid} is computed by evaluating the Fr\'echet distance \citep{dowson1982frechet} between the penultimate network activations computed on both the true and synthetic images.  Today, comparing high-level InceptionV3 features using an \ac{fid}-score \citep{heusel2017gans} enjoys widespread adoption and several variants exist. \acf{kid} \citep{binkowski18kid}, for example, relaxes the multi-variate Gaussian assumption of \ac{fid} and measures the polynomial kernel distance between Inception features of the generated and the training dataset. \cite{binkowski18kid} kept the InceptionV3 backbone and replaced the Fr\'echet distance with a kernel distance. 
While \ac{fid} captures general trends well, the literature also discusses its drawbacks.
\cite{Kynkaat2023TheRole} empirically studied the effect of ImageNet classes on \ac{fid} for non-ImageNet datasets by using GradCAM. Furthermore, \cite{Kynkaat2023TheRole} examined ImageNet bias using \ac{projgan} and StyleGAN2. Compared to StyleGAN2, \ac{projgan} produces more accidental distortions like floating heads and artifacts \citep{Sauer2021NEURIPS}. Surprisingly, \ac{projgan}'s \ac{fid} is comparable to StyleGAN2's in their experiment.
\cite{chong20fidbash} found a generator-dependent architecture bias, which limits the ability to compare samples for smaller datasets with 50K or fewer images.
Additionally, \cite{parmar2022aliased} found that both \ac{fid} and \ac{kid} are highly sensitive to resizing and compression.
\cite{Barrat2018ISNote} reported \ac{fid} sensitivity with respect to different InceptionV3 weights.
While comparing Tensorflow and PyTorch implementations, \cite{parmar2022aliased} measured inconsistent scores due to differing resizing implementations.
Finally, \ac{fid} scores are hard to reproduce unless all details regarding its computation are carefully disclosed \citep{HuggingfaceEval}.
\cite{stein2023exposing} proposed an alternative to over-reliance on InceptionV3, by replacing it with the DINOv2-ViT-L/14 model~\citep{oquab2023dinov2}.
This replacement partially addresses the domain bias problem but significantly increases computational cost. Unfortunately, DINOv2's training dataset is not publicly available.
Furthermore, existing frequency-based metrics such as \ac{swd} proposed in \cite{karras2018progGAN} involves multiple projections on a random basis. In spite of its ability to detect domain bias, it suffers from reproducibility issues \citep{nguyen2023markovian} due to random projections.  Consequently, gaps in the dataset remain hidden. This situation motivates the search for additional quality metrics. A detailed discussion of spectral methods and generative architectures is presented in supplementary Section~\ref{sec:extended_related_work}.

\section{Fr\'echet Wavelet Distance (FWD)}\label{sec:fwd}
\begin{figure}
  \centering
  \includegraphics[scale=0.35]{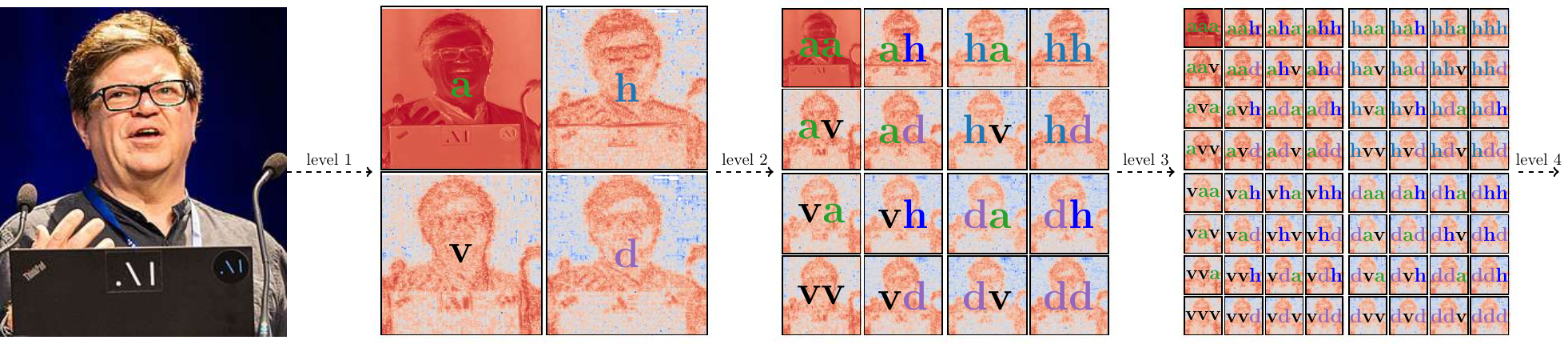}
  \caption{Illustration of the \acf{wpt}. For visualization purposes, we depict a level-3 transform. All later experiments use a level-4 transform. Image from \cite{WikiCun}.}
  \label{fig:wpt_lecun}
\end{figure}
We want to tackle the problem of dataset-domain bias. To this end, we propose \ac{fwd}, which in turn leverages the \acf{wpt}.
We require two-dimensional filters for image processing. We start with single-dimensional Haar wavelets. Next, we construct filter quadruples from the original single-dimensional filter pairs. The process uses outer products \citep{vyas2018multiscale}:
\begin{align}\label{eq:2dfilters}
\mathbf{h}_{a} = \mathbf{h}_\mathcal{L}\mathbf{h}_\mathcal{L}^T,
\mathbf{h}_{h} = \mathbf{h}_\mathcal{L}\mathbf{h}_\mathcal{H}^T,
\mathbf{h}_{v} = \mathbf{h}_\mathcal{H}\mathbf{h}_\mathcal{L}^T,
\mathbf{h}_{d} = \mathbf{h}_\mathcal{H}\mathbf{h}_\mathcal{H}^T,
\end{align}
with $a$ for the approximation filter, $h$ for the horizontal filter, $v$ for the vertical filter, and $d$ for the diagonal filter \citep{lee2019pywavelets}.
We construct a \ac{wpt}-tree for images with these two-dimensional filters, as illustrated
in Fig.~\ref{fig:wpt_lecun}, using recursive convolution operations with the filter quadruples, i.e.,
\begin{align}\label{eq:conv_wpt}
  \mathbf{C}_{\mathcal{F}_{l}} * \mathbf{h}_j = \mathbf{C}_{\mathcal{F}_{l+1}},
\end{align}
at every recursion step where $*$ denotes a two-dimensional convolution with a stride of two. The filter codes $\mathcal{F}_{l+1}$ are constructed by applying all $j \in [a,h,v,d]$ filters to the previous filter codes $\mathcal{F}_{l}$. Initially, the set of inputs $\mathit{F}_{l}$ will only contain the original image $\mathbf{C}_{\mathit{F}_{0}} = \{ X \}$ as shown in Fig.~\ref{fig:wpt_lecun}. 
At level one, we obtain the result of all four convolutions with the input image and have $\mathit{F}_{1} =  [a,h,v,d]$. At level two, we repeat the process for all elements in $\mathit{F}_{1}$. $\mathit{F}_{2}$ now contains two-character keys $[aa, ah, av, ad, \dots, dv, dd]$ as illustrated in Fig.~\ref{fig:wpt_lecun}. We typically continue this process until level 4 in this paper. We arrange the coefficients in $\mathbf{C}_{\mathcal{F}_{l}}$ as tensors $\mathbf{C}_l \in \mathbb{R}^{P,H_p,W_p}$ for the final layer. The total number of packages at every level is given by $P = 4^l$, $H_p = \frac{H}{4^l}$ and $W_p = \frac{W}{4^l}$, where we denote the image height and width as $H$ and $W$.
We provide more details on \ac{wpt} in the Supplementary.

\begin{figure}
  \centering
  \includegraphics[width=.8\linewidth]{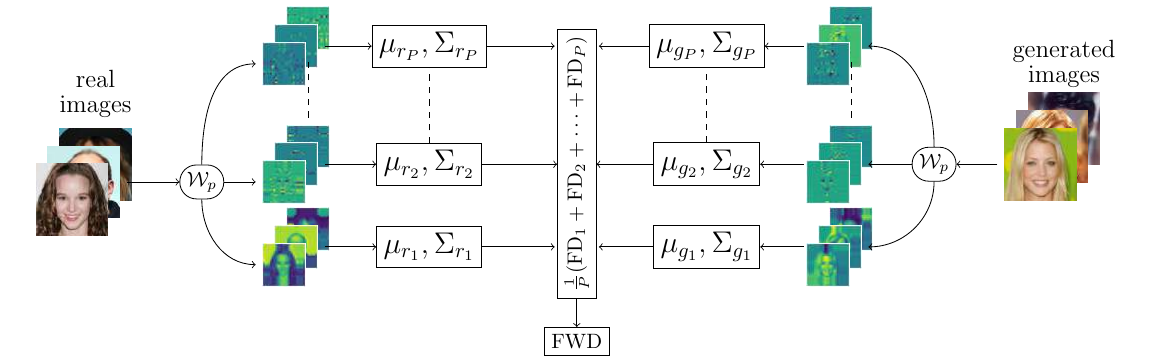}
  \caption{\acf{fwd} computation flow-chart. \acs{wpt} denotes the wavelet-packet transform. Not all packet coefficients are shown, dashed lines indicate omissions. We compute individual Fr\'echet Distances for each packet coefficient and finally average across all the coefficients.}
    \label{fig:fwd_formulation}
\end{figure}

Figure~\ref{fig:fwd_formulation} illustrates how we compute the \ac{fwd}. The process relies on the wavelet packet transform, as previously discussed. We process $N$ images with $C$ channels in parallel $\mathcal{W}_p : I_s\in\mathbb{R}^{N\times H\times W\times C} \rightarrow \mathbf{C} \in \mathbb{R}^{N\times P\times H_p \cdot W_p \cdot C}$, where $H$ and $W$ denote image height and width as before.
To facilitate the ensuing metric evaluation, we flatten the last axes into $(H_p \cdot W_p \cdot C)$. Before computing the packets, all pixels are divided by 255 to re-scale all values to [0,1]. 
The metric is computed in three steps. First, we compute the individual packet mean via
\begin{align}
  \mu_{p}(I_N) = \frac{1}{N}\sum_{n=1}^{N} \mathcal{W}(I_n)_{p},
\end{align}
where $I_n$ is the $n^{th}$ image in the dataset and $p$ represent the corresponding packet from $P$ packets.
Then we compute the covariance matrix as
\begin{align}
  \Sigma_{p}(I_N) = \frac{1}{N-1}\sum_{n=1}^{N}(\mathcal{W}(I_n)_p - \mu_p(I_N))(\mathcal{W}(I_N)_p - \mu_p(I_N))^T.
\end{align}
Here, $\mu \in \mathbb{R}^{P \times C \cdot H_p \cdot W_p}$ denotes the mean across the number of images, and $\Sigma \in \mathbb{R}^{P \times C \cdot H_p \cdot W_p  \times C \cdot H_p \cdot W_p}$ represents the covariance among all the coefficients.
Now we are ready to compute the distances given the packet mean and covariance values,
\begin{equation} \label{eq:frechet_distance}
    \text{FD}_p(r, g) = d(\mathcal{N}(\mu_{r_{p}}, \Sigma_{r_{p}}), \mathcal{N}(\mu_{g_{p}}, \Sigma_{g_{p}}))^2 = ||\mu_{r_{p}} - \mu_{g_{p}}||_2^2 + \text{Tr}[\Sigma_{r_{p}} + \Sigma_{g_{p}} - 2\sqrt{\Sigma_{r_{p}}\Sigma_{g_{p}}}],
\end{equation}
with $r$ and $g$ denoting the real and generated images and $\text{Tr}$ denoting the trace operation. Utilising the above computed per-packet statistics for both real $(\mu_r, \Sigma_g)$ and generated samples $(\mu_r, \Sigma_g)$, we measure the mean of Fr\'echet Distance (Equation~\ref{eq:frechet_distance}) across all packets
\begin{align}
  \text{FWD} = \frac{1}{P}\sum_{p=1}^{P}d(\mathcal{N}(\mu_{r_{p}}, \Sigma_{r_{p}}), \mathcal{N}(\mu_{g_{p}}, \Sigma_{g_{p}}))^2.
\end{align}
By averaging the distances of all frequency bands, the \ac{fwd} captures frequency information across the spectrum.

\section{Experiments}\label{sec:experimenta}
Our first series of experiments demonstrates the effect of domain bias on learned metrics, demonstrating the resilience of \ac{fwd} to such bias.
All experiments were implemented using the same code base. \\
\textbf{Implementation:} We use PyTorch~\citep{paszke2017automatic} for neural network training and evaluation and compute \ac{fid} using \citep{Seitzer2020FID} as recommended by \cite{Heusel2017FidCode}.
We work with the wavelet filter coefficients provided by PyWavelets~\citep{lee2019pywavelets}. We chose the PyTorch-Wavelet-Toolbox~\citep{ptwt2024JMLR} software package for GPU support. 
\acs{fd_dino} and \ac{kid} are computed using the codebases from \cite{stein2023exposing} and \cite{binkowski18kid}, respectively.


\subsection{Effect of domain bias}\label{sec:domain_bais}

\begin{table}
  \centering
  \caption{Comparison of \ac{fid}, \ac{fd_dino} and \ac{fwd} to depict domain bias. \ac{fid} prefers \ac{projgan} over \ac{ddgan} across all the datasets. Whereas \ac{fwd} prefers \ac{ddgan}. We find that \ac{fd_dino} agrees with \ac{fwd} across all datasets except \ac{agriculture}. This might be because agriculture data is not part of DINOv2's training set.}
  \label{tab:inet_bias_values}
  \begin{tabular}{ccccc}
  \toprule
  Dataset   & Generator         & \acs{fid}$\downarrow$ & \acs{fd_dino}$\downarrow$ & \acs{fwd}$\downarrow$ (ours) \\ \midrule 
  \multirow{2}{*}{
    \rotatebox[origin=c]{0}{\acs{celebahq}}
  }
            & \acs{projgan}     & \textbf{6.358} & 685.889 & 1.388         \\ 
            & \acs{ddgan}       & 7.641          & \textbf{199.761} & \textbf{0.408}\\ \midrule 
  \multirow{2}{*}{
    \rotatebox[origin=c]{0}{\acs{ffhq}}
  }
            & \acs{projgan}     & \textbf{4.106} & 593.124 &  0.651         \\ 
            & StyleGAN2         & 4.282 & \textbf{420.273} & \textbf{0.312} \\ \midrule
            \multirow{2}{*}{
    \rotatebox[origin=c]{0}{\acs{agriculture}}
  }
            & \acs{projgan}     & \textbf{4.675} & \textbf{171.625} & 1.442         \\ 
            & \acs{ddgan}       & 26.233         & 232.884 & \textbf{1.357}\\  \midrule 
  \multirow{2}{*}{
    \rotatebox[origin=c]{0}{Sentinel}
  }
            & \acs{projgan}     & \textbf{8.96} & 424.898 & 0.755 \\ 
            & \acs{ddgan}       & 23.615        & \textbf{404.700} & \textbf{0.115}\\  \bottomrule 
\end{tabular}
\end{table}
\begin{figure}
  \centering
  \begin{subfigure}[b]{0.45\textwidth}
  \centering
  \includegraphics[width=.9\textwidth]{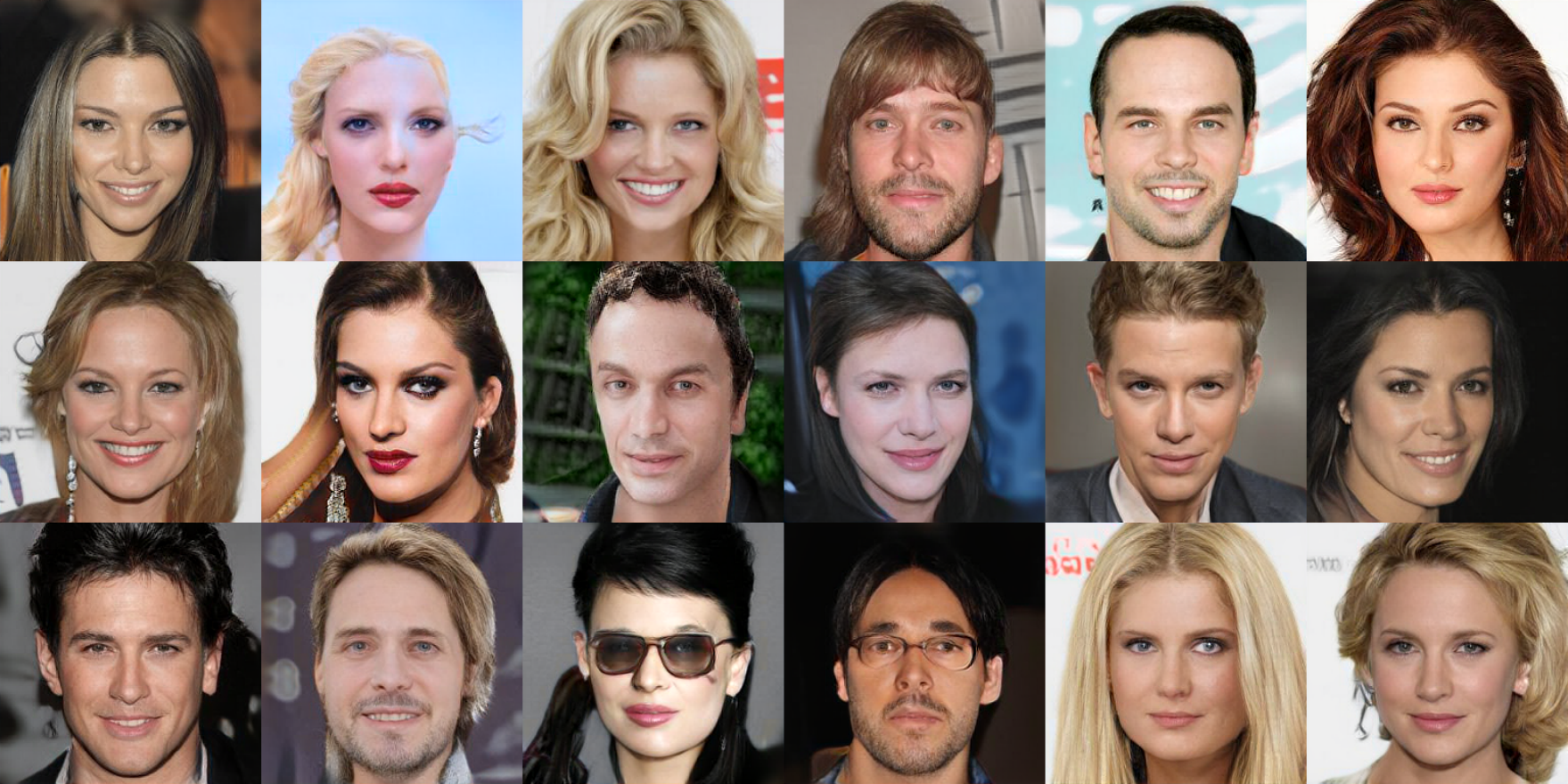}
  \caption{\acs{projgan} on \acs{celebahq} (\acs{fid}: 6.358, \acs{fwd}: 1.388)}
  \label{fig:pggan_celeba_bias}
  \end{subfigure}
  \hspace{1cm}
  \begin{subfigure}[b]{0.45\textwidth}
  \centering
  \includegraphics[width=.9\textwidth]{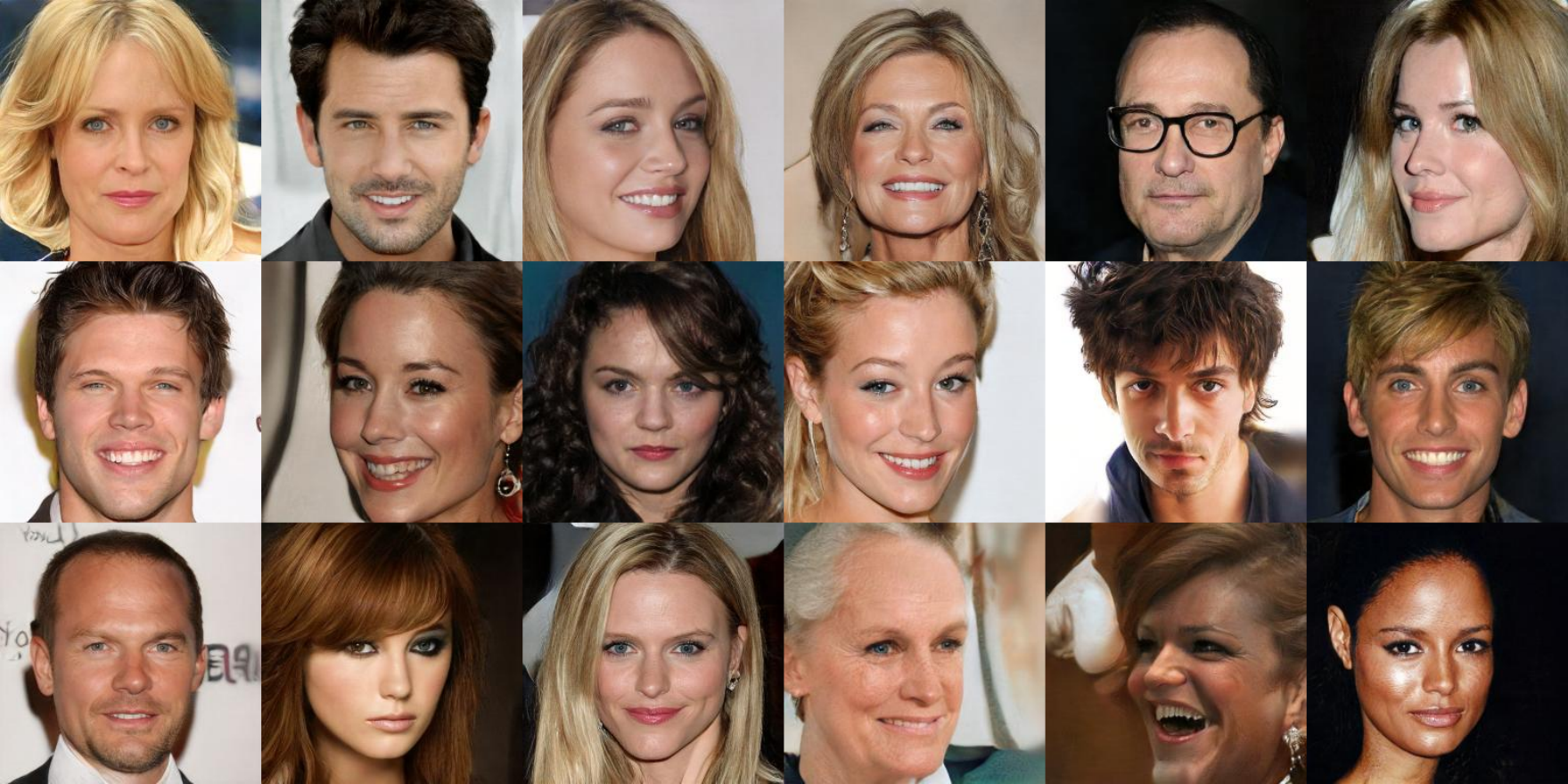}
  \caption{\acs{ddgan} on \acs{celebahq} (\acs{fid}: 7.641, \acs{fwd}: 0.408)}
  \label{fig:ddgan_celeba_bias}
  \end{subfigure}
  \caption{Samples from (a) \ac{projgan} and (b) \acs{ddgan} on the \ac{celebahq} dataset.
  The \ac{fid} prefers \acs{projgan} irrespective of visual artefacts and floating heads, whereas our metric (\ac{fwd}) ranks \acs{ddgan} higher than \ac{projgan}.}
\end{figure}

\cite{Kynkaat2023TheRole} observed that metrics based on ImageNet-trained network features emphasize ImageNet-related information.
This behaviour is desired when we evaluate generators on ImageNet or similar datasets. When working with other datasets, this behaviour is misleading. \\
\textbf{Datasets:} As datasets, we use \ac{celebahq} \citep{karras2018progGAN}, \ac{ffhq}, \acs{agriculture} \citep{yi2020agridata}, an agricultural dataset, and Sentinel \citep{schmitt2019sen12ms}, a remote sensing dataset. These datasets contain images that are very different from those in ImageNet.  More information about the \ac{agriculture} and the Sentinel dataset can be found in the supplementary material.\\
\textbf{Generators:} We study dataset domain bias effects using the \ac{ddgan}, \ac{projgan} and StyleGAN2 networks. \ac{projgan} is particularly interesting since its discriminator relies on ImageNet weights to improve training convergence \citep{Sauer2021NEURIPS}.
Prior work found this architecture to improve \ac{fid} on image datasets far from ImageNet, without substantially improving image quality \citep{Kynkaat2023TheRole}.  \\
\textbf{Hyperparameters:} To examine the effect of dataset bias, we require generators, which are tuned to produce output that resembles our datasets' distribution.
Specifically, we trained \ac{projgan} for 100 epochs on both the \ac{celebahq} dataset and \ac{agriculture}, respectively, using a learning rate of 1e-4 and batch size of 64 with 8 A100 GPUs.
For the Sentinel dataset, we trained \ac{projgan} for 150 epochs, using the same hardware and hyperparameters.
For FFHQ, pre-trained weights are available, as well as pre-trained weights for \ac{ddgan} on \ac{celebahq} from  \cite{xiao2022tackling}.
On \ac{agriculture}, we trained \ac{ddgan} for 150 epochs with a learning rate of 1e-4 and batch size of 8 on the same hardware.
We also trained \ac{ddgan} on the Sentinel dataset for 250 epochs, using a learning rate of 1e-4 and batch size of 4 on 4 A100 GPUs.
For StyleGAN2, we use the pretrained weights with the code from \cite{karras2020analyzing}.  \\

\begin{figure}
  \centering
  \begin{subfigure}[b]{0.4\textwidth}
    \centering
    \includestandalone[width=\textwidth]{./figures/histograms/celeba_hist}    
    \caption{\acs{celebahq}}
    \label{fig:hist_celebahq_bias}
  \end{subfigure}
  \begin{subfigure}[b]{0.4\textwidth}
    \centering
    \includestandalone[width=\textwidth]{./figures/histograms/agri_data_hist}
    \caption{\acs{agriculture}}
    \label{fig:hist_agri_bias}
  \end{subfigure}
  \caption{Distribution of ImageNet Top-1 classes, predicted by InceptionV3 for real images and images generated by \ac{ddgan} and \ac{projgan}. (a) depicts the distribution for the \ac{celebahq} dataset and (b) shows the distribution for \ac{agriculture}. Although irrelevant for visual quality, the class distribution of \ac{projgan} aligns more closely with the real distribution than \ac{ddgan} for both the datasets, contributing to lower \ac{fid} for \ac{projgan}.}
  \label{fig:top15-agri}
\end{figure}

\textbf{Results:} Table~\ref{tab:inet_bias_values} presents the \ac{fid}, \ac{kid}, \ac{fd_dino} and \ac{fwd} values across all datasets for the aforementioned generators.  Across all datasets, \ac{fid} prefers images generated by \ac{projgan}. 
When we compare images generated by \ac{projgan} and \acs{ddgan} for \acs{celebahq}, which are shown in Figures~\ref{fig:pggan_celeba_bias} and \ref{fig:ddgan_celeba_bias}, we observe that more deformations are visible in the images of \ac{projgan} compared to \acs{ddgan} images. \acs{ddgan}, in other words, produces more high-quality images. Supplementary Figures \ref{fig:ddgan_celebahq_full} and \ref{fig:pggan_celebahq_full} illustrate this observation further.
Consequently, it is surprising to see \ac{fid} preferring \ac{projgan}, as we would expect \ac{ddgan} to come out on top.
Following \cite{Kynkaat2023TheRole}, we compare the InceptionV3 output label distribution of the original-\ac{celebahq} images to their synthetic counterparts from \ac{ddgan} and \ac{projgan} in Figure~\ref{fig:hist_celebahq_bias}. 
We observe that InveptionV3 produces a label distribution for \ac*{projgan}, which resembles the distribution from InveptionV3 for the original \ac*{celebahq} images. The label distribution for images from \ac{ddgan} differs significantly. This discrepancy, also reported by \cite{Kynkaat2023TheRole},
explains why \ac{fid} produces a misleading verdict. \ac{fwd}, in contrast, prefers \ac{ddgan}, as we would expect.

The same pattern repeats in the results for our \ac{ffhq}-experiments. Generally, we see \ac{fid} preferring \ac{projgan} images, while \ac{fwd} puts StyleGAN2 on top. Our observations confirm the experiment in \cite{Kynkaat2023TheRole}.
In a next step, we study the effect of a larger network backbone for the neural Fr\'echet distance computations. \cite{stein2023exposing} proposed to replace InveptionV3 with the much larger pretrained DINOv2 network. Table~\ref{tab:inet_bias_values} lists the resulting distance metrics. For \ac{celebahq} and \ac{ffhq}, \acs{fd_dino} prefers \ac{ddgan} and StyleGAN2 images respectively. Here, \acs{fd_dino} and \ac{fwd} agree.

To investigate further, we consider the \ac{agriculture} of agricultural images \citep{yi2020agridata} and the Sentinel \citep{schmitt2019sen12ms} dataset. Samples from \ac{projgan} for \ac{agriculture} and Sentinel are provided in Figures \ref{fig:pggan_dndd_full} and \ref{fig:pggan_sent_full}, respectively. Correspondingly, Figures \ref{fig:ddgan_dndd_full} and \ref{fig:ddgan_sent_full} represent samples from \ac{ddgan} for \ac{agriculture} and Sentinel, respectively. In both cases, \ac{fid} consistently prefers \ac{projgan}, which was also the case in all prior experiments. Histograms of the InceptionV3 label distribution are
depicted in Figure~\ref*{fig:hist_agri_bias}. The histograms indicate domain bias and resemble the observations reported above. 
On \ac{agriculture} and Sentinel, the verdicts of \acs{fd_dino} and \acs{fwd} are particularly interesting. While both metrics correctly agree on the Sentinel dataset, only \ac{fwd} correctly prefers \ac{ddgan} on the agricultural images.

We carefully chose the \acs{agriculture}, as agriculture images are not commonly used, and the dataset does not resemble ImageNet. We speculate that the LVD-142M dataset may include satellite imagery, contributing to a consistent ranking. Unfortunately, the closed source of the LVD-142M dataset used for training DINOv2 \citep{oquab2023dinov2} makes it difficult to investigate this domain bias more in detail.
In this first set of experiments, we observed that, while \ac{fd_dino} provides a partial remedy to the domain bias problems, it still produces an inconsistent ordering for the \ac{agriculture} images. Furthermore, this partial remedy comes at a tremendous computational cost. Table~\ref{tab:eff_comparison} shows that \ac{fwd} is over 36 times faster to compute than \ac{fd_dino}.

\begin{table}
  \centering
  \caption{Comparison of computational efficiency between \ac{fid}, \ac{fd_dino} and \ac{fwd}. \ac{fwd} exhibits the lowest FLOPs and highest throughput. \ac{fd_dino} has the highest FLOPs and lowest throughput because of its large network, and \ac{fid} is in between. FLOPs are calculated over individual feature extractors on a single image, and throughput is measured over 50k images.}
  \label{tab:eff_comparison}
  \resizebox{0.45\textwidth}{!}{
  \begin{tabular}{ccc}
    \toprule
    Metric & GFLOPs$\downarrow$ & Throughput (imgs/sec)$\uparrow$ \\ \midrule
    \acs{fid}    &   1.114    &    526       \\
    \acs{fd_dino}&   15.566   &     53       \\
    \acs{fwd}    &   \textbf{0.006}    &    \textbf{1923}      \\ \bottomrule
  \end{tabular}}
\end{table}
\begin{table}
  \centering
  \caption{Evaluation of \ac{fid} (ImageNet), \ac{fid} (CelebA) and \ac{fwd} on the \ac{celebahq} and \ac{ffhq} datasets. \ac{fid} (ImageNet) prefers \ac{projgan} in both datasets, whereas \ac{fid} retrained on CelebA and \ac{fwd} both prefer \ac{ddgan} in these datasets.}
  \label{tab:fid_celebahq}
  \resizebox{0.83\textwidth}{!}{
  \begin{tabular}{ccccc}
    \toprule
    Dataset & Generator & \acs{fid} (ImageNet)$\downarrow$ & \acs{fid} (CelebA)$\downarrow$ & \acs{fwd}$\downarrow$ \\ \midrule
    \multirow{2}{*}{\acs{celebahq}} & \acs{projgan} & \textbf{6.358} & 5.602 & 1.388 \\
                                    & \acs{ddgan}   & 7.641 & \textbf{3.145} & \textbf{0.408} \\ \midrule
    \multirow{2}{*}{\acs{ffhq}}     & \acs{projgan} & \textbf{4.106} & 2.204 & 0.651 \\
                                    & StyleGAN2   & 4.282 & \textbf{0.897} & \textbf{0.312} \\ \bottomrule
  \end{tabular}}
\end{table}

In a second series of experiments, we investigate the effect of retraining, another expensive solution to the domain bias problem. To this end, we train InceptionV3 on \ac{celeba}.
\ac{celeba} comes with 40 facial attributes, which we use to train a classifier. After convergence, we see an exact match ratio of 90\% and recalculate \ac{fid} using this new backbone. The \ac{fid} (CelebA) column of Table~\ref{tab:fid_celebahq} lists the corresponding scores, and \ac{fid} (CelebA) and \ac{fwd} provide the same order.

However, in the case of the agricultural dataset, the retrained \ac{fid} (DNDD) in supplementary Table~\ref{tab:fid_dndd} remains biased, while \ac{fwd} produces meaningful domain agnostic results. \ac{agriculture} contains 3600 images with seven classes and the task requires detecting nutrient deficiency in winter wheat and winter rye, such as nitrogen, phosphorous, and potassium deficiencies. Once more, we use a retrained InceptionV3 backbone for the \ac{fid} computation.  Compared to \ac{celeba} or ImageNet, this is a small dataset and the retrained network does not provide meaningful features.  
This is an interesting use case since it illustrates that \ac{fwd} is not just free from data bias. It also provides meaningful feedback for low-resource tasks where retraining InceptionV3 is not feasible.


In conclusion, experiments in this section indicate that metrics like \ac{fid} and \ac{fd_dino}, while helpful, are prone to domain bias when applied to datasets beyond the underlying training datasets.
On the contrary, \ac{fwd} offers a computationally efficient, consistent and domain-agnostic evaluation. 


\subsection{\ac{fwd} interpretability}\label{sec:interpertabilitatio}

\begin{figure}
  \centering
  \begin{subfigure}[b]{0.28\textwidth}
    \centering
    \includestandalone[width=0.93\textwidth]{./figures/packet_transform/lvl_3_packs_new}
    \caption{Filter order}
  \end{subfigure}
  \hfill
  \begin{subfigure}[b]{0.28\textwidth}
    \centering
    \includegraphics[width=0.94\textwidth]{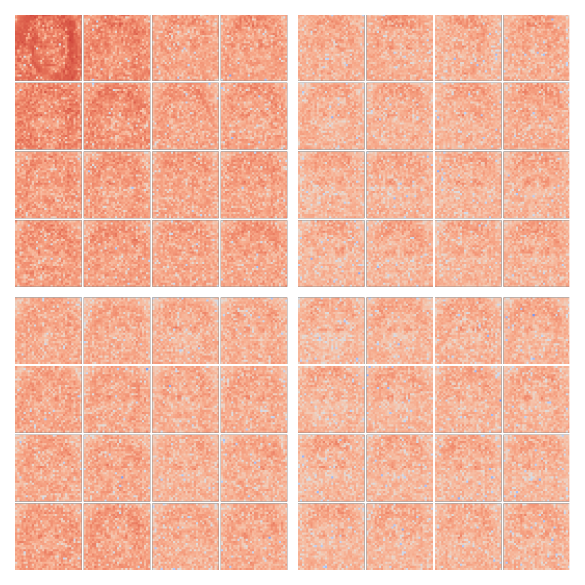}
    \caption{StyleGAN2}
    \label{fig:fwd_interpre_sgan2}
  \end{subfigure}
  \hfill
  \begin{subfigure}[b]{0.28\textwidth}
    \centering
    \includegraphics[width=0.94\textwidth]{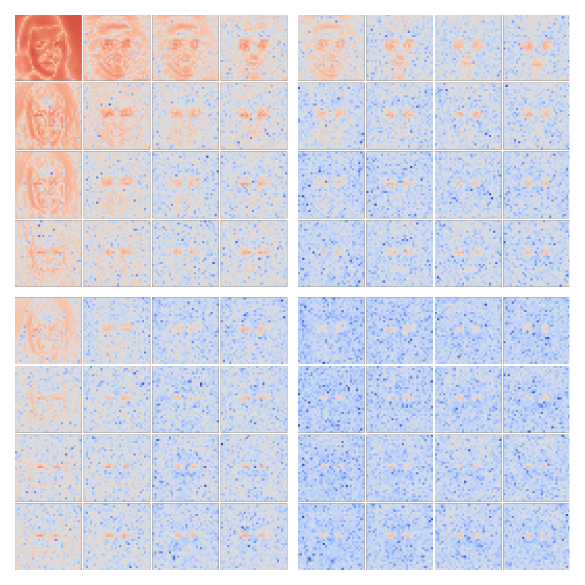}
    \caption{DDGAN}
    \label{fig:fwd_interpre_ddgan}
  \end{subfigure}
  \hfill
  \begin{subfigure}[b]{.1\textwidth}
    \centering
    \includegraphics[width=0.88\textwidth]{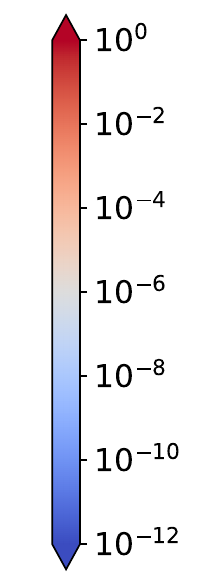}
    \caption*{}
  \end{subfigure}
  \hfill
  \begin{subfigure}[b]{\textwidth}
    \centering
    \includegraphics[scale=0.25]{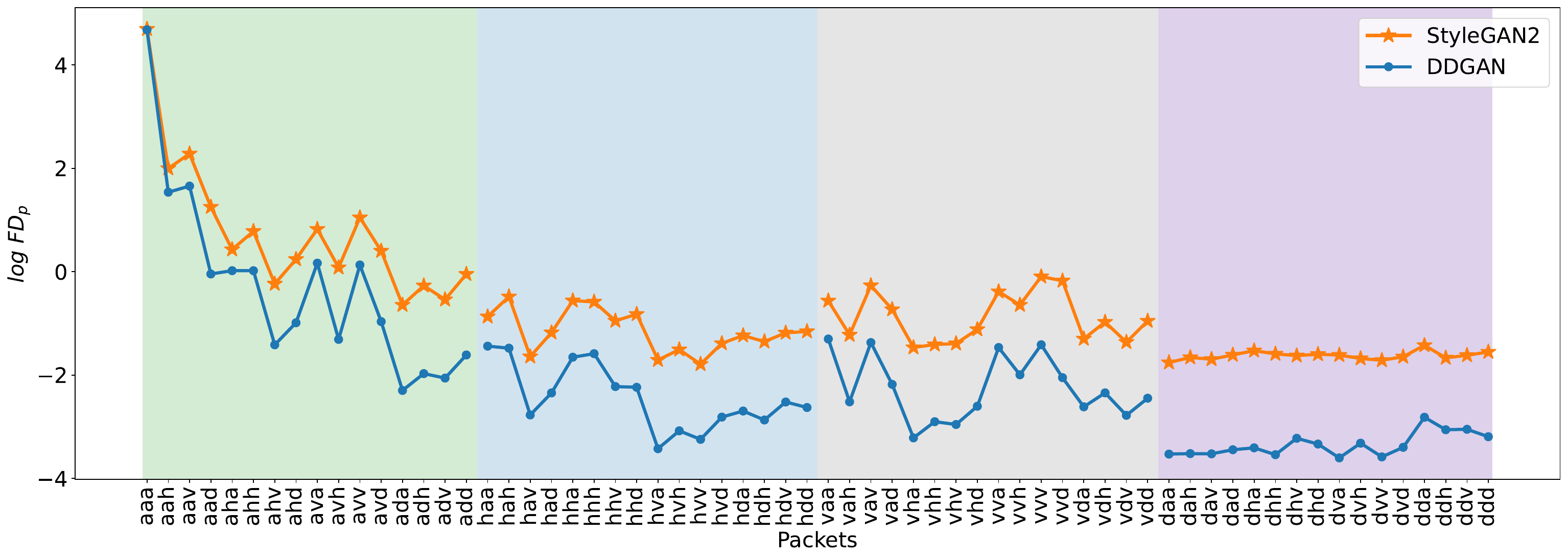}
    \caption{}
    \label{fig:log_fd_pro_packt}
  \end{subfigure}
  \caption{Interpretation of FWD. (a) represents the blueprint for level-3 \acs{wpt} transformation. (b) and (c) depict the mean absolute packet difference between \ac{celebahq} dataset and generated images by StyleGAN2 and \acs{ddgan}, respectively. (d) shows the per-packet Fr\'echet distances for StyleGAN2 in orange and DDGAN in blue.
  }
  \label{fig:fwd_interpre}
\end{figure}

A generative metric is interpretable if and only if we can understand the underlying mechanics that produce the ranking. This section explains the decisions made by \ac{fwd} in one specific case where we focus on samples from \ac{ddgan} and StyleGAN2 for \ac{celebahq}.

Section~\ref{sec:fwd} formulates \ac{fwd} as an average of per packet \ac{fwd} scores.
This design choice allows us to understand the overall \ac{fwd}-score in terms of the individual packet coefficients for each frequency band.
Figures~\ref{fig:fwd_interpre_sgan2} and \ref{fig:fwd_interpre_ddgan} depict the mean absolute difference per packet between the original images of \ac{celebahq} and generated samples from StyleGAN2 and \ac{ddgan}, respectively.
Figure~\ref{fig:log_fd_pro_packt} presents both generators' per-packet \ac{fwd}.
Figure~\ref{fig:fwd_interpre}d shows that \ac{ddgan} has a lower Fr\'echet distance for all packets and averaging the distances over all packages translates into a meaningful metric.    

\subsection{Evaluation of Robustness}\label{sec:robust}

The section follows up on prior work by \cite{Kynkaat2023TheRole}.
The authors generate a large set of samples and find a specific combination of images with an optimal \ac*{fid}. First, the weights of each image are optimized with \ac{fid} as the objective function. Second, a subset of images is sampled based on the weights.  
We follow this process and sample 50k images from a large set with optimized weights as probabilities. We employ generated images from StyleGAN2 and real-world images from the \ac{ffhq} dataset.
Supplementary Table~\ref{tab:fid_opt} lists the resulting \ac{fid}, \ac*{fd_dino} and \ac{fwd} values. We observe that \ac{fwd} is robust to \ac{fid} optimization, whereas \ac*{fd_dino} showed a little reduction by optimizing \ac*{fid}. 

In addition to \ac{fid} optimization, we study the impact of image perturbation in supplementary Figure~\ref{fig:corruption_robustness}. We find that \ac{fwd} and \ac*{fd_dino} are closer to a bijective mapping in the presence of perturbation than \ac{fid}. This behaviour is desirable since we would always expect a larger distance if for example more noise is added. This is not always the case for \ac{fid}. Consider for example the last quarter of the uniform noise intensity in (b), where \ac{fid} falls even though more noise is added.






\subsection{Comparison to State of the art}
\begin{table}
      \centering
      \caption{Comparing various generative models using \acf{fwd}, \acf{fid}, \acf{fd_dino} and \acf{kid} on the \ac{celebahq}, LSUN-Churches, LSUN-Bedrooms and ImageNet datasets.}
      \label{tab:new_metrics}
      \resizebox{\textwidth}{!}{
      \begin{tabular}{c c c c c c c}
      \toprule
      Dataset             & Image Size          & Method                        & \acs{fid}$\downarrow$ & \ac{kid}$\downarrow$ & \acs{fd_dino}$\downarrow$ & \acs{fwd}$\downarrow$ (ours) \\ \midrule
      \multirow{5}{*}{
        \rotatebox[origin=c]{90}{CelebAHQ}
      }       & \multirow{5}{*}{256}                      & \acs{ddim}~\cite{song2021ddim}&   32.333              & 0.0313           &  654.482    & 12.317\\
      & & \acs{ddpm}~\citep{ho20ddpm}                                  &   19.101              & 0.0152                &       341.838       & 4.697\\ 
      & & StyleSwin~\citep{zhang2022styleswin}                         &   23.257              & 0.0264                &       255.404       & 1.528\\                             
      & & StyleGAN2~\citep{karras2020analyzing}                        &   15.439              & 0.0155                &       593.344       & 0.476\\   
      & & \acs{ddgan}~\citep{xiao2022tackling}                          &   \textbf{7.203}              & \textbf{0.0034}   &  \textbf{199.761}       & \textbf{0.408}\\ \midrule
     \multirow{4}{*}{
      \rotatebox[origin=c]{90}{Churches}   
     } & \multirow{4}{*}{256}
     & \acs{ddim}~\citep{song2021ddim}                               & 11.775                & 0.0043                  &    538.400    & 4.919\\ 
     & & \acs{ddpm}~\citep{ho20ddpm}                                   & 9.484                 & 0.0036                &    454.402    & 3.546\\
     & & StyleSwin~\citep{zhang2022styleswin}                          & \textbf{3.187}        & \textbf{0.0005}       &    \textbf{435.967}    & 2.835 \\                                                             
     & & StyleGAN2~\citep{karras2020analyzing}                         & 4.309                 & 0.0007                &    444.044    & \textbf{0.753}\\ \midrule 
     \multirow{4}{*}{
      \rotatebox[origin=c]{90}{Bedrooms}
     }     & \multirow{4}{*}{256}                         & & &\\ 
     & & \acs{ddim}~\citep{song2021ddim}                   & 25.857          & 0.0094          & 452.419 &  9.521\\
     & & \acs{ddpm}~\citep{ho20ddpm}                       & \textbf{16.251} & \textbf{0.0058} & \textbf{392.481} & \textbf{5.187}\\
                                    & & & &\\                              
                                    \midrule
      \multirow{4}{*}{
        \rotatebox[origin=c]{90}{ImageNet}}
            &\multirow{4}{*}{64}                    & Imp. Diff. (VLB)~\citep{nichol2021improvdiff}       & 33.522        & 0.0264               & 670.952 & 2.182 \\       
            &                       & EDM~\citep{karras2024analyzing}                                             & 12.295        & 0.0108               & 113.704 & 1.160\\ 
            &                       & BigGAN~\citep{brock2018biggan}                                             &  5.128        & 0.0024               & 170.601 & 0.441 \\
            &                       & Imp. Diff. (Hybrid)~\citep{nichol2021improvdiff}    & \textbf{3.091}& \textbf{0.0006}      & \textbf{96.208}  & \textbf{0.392}\\\bottomrule
    \end{tabular}}
    \vspace{-0.5cm}
\end{table}
To understand the spectral qualities of existing generative methods for image synthesis, we evaluated various Diffusion and \ac{gan} models across a wide range of benchmark datasets. \\
\textbf{Datasets:} We compare common metrics and our \ac{fwd} on \ac{celebahq} \citep{karras2018progGAN}, the Church and Bedroom subsets of the \ac{lsun} dataset \citep{Yu2015LSUNCO}, and finally ImageNet \citep{imagenet15russakovsky}. In order to retain consistent spatial and frequency characteristics across various image sizes, we use the level 4 packet transform for 256x256 images. For images that are smaller, we use fewer levels, i.e., 3 for 128x128 and 2 for 64x64.    
\textbf{Generators:} For the evaluation, we use the diffusion approaches \ac{ddpm} \citep{ho20ddpm}, \ac{ddim} \citep{song2021ddim}, Improved Diffusion \citep{nichol2021improvdiff}, DDGAN \citep{xiao2022tackling}, EDM \citep{karras2024analyzing}, as well as the GAN approaches StyleGAN2 \citep{karras2019style}, StyleSwin \citep{zhang2022styleswin} and BIGGAN \citep{brock2018biggan}.\\
\textbf{Hyperparameters:} All generators are evaluated with pretrained weights as provided by the respective paper codebases. \\
\textbf{Metrics:} Table~\ref{tab:new_metrics} reports the results for \ac{fid}, \ac{kid}, \ac{fd_dino} and finally our own \ac{fwd}. \ac{fid}-scores are obtained by the standard implementation by \cite{Seitzer2020FID}. 
The ImageNet numbers are computed with 50k images from the validation set. For CelebAHQ and LSUN, we work with 30k images.

Considering \ac{celebahq}, \ac{fid}, \ac{kid}, \ac{fd_dino} and \ac{fwd} agree most of the time. Considering \ac{fid} and \ac{fwd}, only \acs{ddpm} and StyleSwin are swapped. It is interesting to note that \ac{fwd} ranks \acs{ddim}, \acs{ddpm}, StyleSwin, and StyleGAN2 on \ac{celebahq} and LSUN churches the same way, whereas \ac{fid} ranks StyleSwin differently on the two datasets. According to \ac{fid}, StyleSwin performs worse than \acs{ddpm} and StyleGAN2 on \ac{celebahq} but better than these two approaches on LSUN churches. This is counterintuitive, but it can be explained by the domain bias of \ac{fid}. The supplementary Figure~\ref{fig:celebahq_hist} depicts the histograms of top-1 classes classified by InceptionV3 on \ac{celebahq} for \ac{ddpm} and StyleSwin.
We observe that \ac{ddpm} matches the activation histograms of \ac{celebahq} more accurately than the histograms of StyleSwin, whereas the histograms of both methods are very similar for LSUN-Church as shown in Figure~\ref{fig:lsun_church_hist}. As a result, \ac{fid} ranks StyleSwin worse on \ac{celebahq} but better on LSUN-Church. Our metric \ac{fwd} is not biased by the class distribution and provides a consistent metric for both datasets.


We also consider the LSUN-Bedrooms and ImageNet 64 datasets, where \ac{fid} and \ac{fwd} agree. We expect pristine performance for \ac{fid} on ImageNet since this setting is perfectly in its data-domain.
Yet, \ac{fd_dino} places EDM \citep{karras2024analyzing} ahead of BigGAN, which is surprising since this does not match with the ranking from \ac{fid}. \ac{fid} and \ac{fwd} agree and arrive at the same ranking.

\subsection{User Study}
\label{sec:user_study}
To ensure that our metric aligns with human perception, we conduct two types of user studies. The first study demonstrates that \ac{fwd} does not suffer from domain bias. The second study supports \ac{fwd}'s alignment with human rankings on large-scale diffusion models. \\
\textbf{Datasets and Generators:} In case of the first user study, we use \ac{celebahq} and \ac{agriculture} to assess the perceptual quality of images generated by \ac{projgan} and \ac{ddgan}. For the second user study, we use Conceptual Captions~\citep{sharma2018conceptual} as the evaluation dataset and work with pre-trained StableDifusion models, particularly versions 1.5, 2.1, 3.0 (Medium), and 3.5 (Large) from hugging face (\url{https://huggingface.co/stabilityai}).

\textbf{Results:}
Table~\ref{tab:domain_bias_human} presents the results of the first user study. A higher \ac{her} in the table implies that the participants find the generated images more realistic than the original images. The \ac{her} results show that the users identify \ac{ddgan} generated images more realistic than \ac{projgan} generated images. Predominantly, this table highlights \ac{fwd}'s alignment with user preferences across both \ac{celebahq} and \ac{agriculture} in comparison to \ac{fid} and \ac{fd_dino}.
Supplementary Table~\ref{tab:stable_diffusion_human} exhibits the overall alignment of \ac{fwd} with human perception on large-scale diffusion models. We observe that \ac{fwd} prefers the latest StableDiffusion-3.5 model over other models, same as the users, whereas \ac{fid} and \ac{fd_dino} rank the StableDiffusion-1.5 model surprisingly better. Moreover, we observe that \ac{fwd} and other metrics prefer StableDiffusion-1.5 images over 2.1 images. On careful observation of images from these models, we observe that the 2.1 model generates images with artifacts like deformed bodies, extra hands, improper artistic images (like paintings), and some images with white contrast more often than the 1.5 model. We provide the samples from all the StableDiffusion models in Supplementary Section~\ref{sec:sd_samples}.
Overall, the user study demonstrates \ac{fwd}'s alignment with human perception and that it does not suffer from  a domain bias.

\begin{table}[]
\centering
\caption{Comparison of existing metrics \ac{fid}, \ac{fd_dino} and \ac{fwd} with \acf{her}. Higher \ac{her} means that participants find the generated images more realistic than the images of the original dataset. \ac{her} shows that \ac{ddgan} generates perceptually better images.}
\label{tab:domain_bias_human}
\resizebox{0.8\textwidth}{!}{
\begin{tabular}{cccccc}
\toprule
   Dataset      &       Generator        & \ac{fid}$\downarrow$    & \ac{fd_dino}$\downarrow$ & \ac{fwd}$\downarrow$   & \acs{her}(\%)$\uparrow$  \\
\midrule
\multirow{2}{*}{\ac{celebahq}} &  \ac{projgan} & \textbf{6.358}  & 685.889   & 1.388 & 20.0 \\
         & \ac{ddgan}         & 7.641  & \textbf{199.761}   & \textbf{0.408} & \textbf{32.5} \\
\midrule
\multirow{2}{*}{\ac{agriculture}} & \ac{projgan} & \textbf{4.675}  & \textbf{171.625}   & 1.442 & 50   \\
         & \ac{ddgan}         & 26.233 & 232.884   & \textbf{1.357} & \textbf{57}   \\
\bottomrule
\end{tabular}}
\vspace{-0.5cm}
\end{table}

\section{Conclusion}\label{sec:conclusion}

Modern generative models exhibit frequency biases \citep{durall2020watch}, while commonly used metrics such as \ac*{fid}, \ac*{kid} and \ac*{fd_dino} are affected by domain bias \citep{Kynkaat2023TheRole}. To address these limitations, \ac{fwd} accounts for frequency information without introducing a domain-specific bias. Even though \ac*{fd_dino} offers a partial solution to this issue, it comes at a very high computational cost and has thus a negative environmental impact.  In response, this paper introduced \ac{fwd}, a novel metric based on the wavelet packet transform. Our metric allows a consistent and domain-agnostic evaluation of generative models, and it is computationally efficient. Our findings show that \ac*{fwd} is robust to input perturbations and interpretable through the analysis of individual frequency bands.
\ac*{fwd} in conjunction with traditional metrics ensures a comprehensive and accurate evaluation of generative models while also helping to mitigate domain bias.

\newpage

\subsubsection*{Acknowledgments}
This research was supported by the Federal Ministry of Education and Research (BMBF) under grant no.\ 01IS22094A WEST-AI and 6DHBK1022 BNTrAInee, the Deutsche Forschungsgemeinschaft (DFG, German Research Foundation) GA 1927/9-1 (KI-FOR 5351) and the ERC Consolidator Grant FORHUE (101044724). Prof. Kuehne is supported by BMBF project STCL - 01IS22067. The authors gratefully acknowledge the Gauss Centre for Supercomputing e.V.\ (www.gauss-centre.eu) for funding this project by providing computing time through the John von Neumann Institute for Computing (NIC) on the GCS Supercomputer JUWELS at Jülich Supercomputing Centre (JSC). The authors heartfully thank all the volunteers who participated in the user study. The sole responsibility for the content of this publication lies with the authors.

\bibliography{iclr2025_conference}
\bibliographystyle{iclr2025_conference}

\newpage

\appendix
\section{Supplementary}

\subsection{Acronyms}
\printacronyms[heading=None]

\subsection{\acs{fwd} Robustness}
To supplement Section~\ref{sec:robust}, we provide results for \ac{fwd}'s robustness towards various perturbations such as Gaussian blur, uniform noise and JPEG compression in Figure~\ref{fig:corruption_robustness}. Furthermore, in Table~\ref{tab:fid_opt}, we demonstrate that matching fringe features can be used to optimise \ac{fid} and \ac{fd_dino}, whereas \ac{fwd} does not improve.
\begin{figure}
  \centering
  \hfill
  \begin{subfigure}[b]{0.32\textwidth}
      \centering
      \includestandalone[width=\textwidth]{./figures/corrupt_plots/blur}
      \caption{}
      \label{fig:blur}
  \end{subfigure}
  \begin{subfigure}[b]{0.32\textwidth}
      \centering
      \includestandalone[width=\textwidth]{./figures/corrupt_plots/uniform}
      \caption{}
      \label{fig:noise_uniform}
  \end{subfigure}
  \hfill
  \begin{subfigure}[b]{0.32\textwidth}
   \centering
   \includestandalone[width=\textwidth]{./figures/corrupt_plots/pil}
   \caption{}
   \label{fig:pil_jpeg}
 \end{subfigure}
  \caption{Figures depicting the effect of perturbations such as (a) Gaussian blur, (b) uniform noise corruption and (c) JPEG compression on \ac{fid}, \ac{fwd} and \ac{fd_dino}.}
     \label{fig:corruption_robustness}
\end{figure}

\begin{table}[]
  \centering
  \caption{Matching fringe features for 250k images generated using StyleGAN2 for the \ac{ffhq} dataset. By optimizing the sample weights for \ac{fid}, \ac{fd_dino} is also slightly improved. In contrast, \ac{fwd} penalizes the manipulation of the sample distribution.  
  }
  \label{tab:fid_opt}
  \begin{tabular}{cccc}
  \toprule
  Metric & Random Images              & FID-Optimized Images          & Change  \\ \hline
  \ac{fid}    & 4.278 $\pm$ 0.019          & \textbf{2.031 $\pm$ 0.005}    &  -52.53\%  \\
  \ac{fd_dino}& 420.223 $\pm$ 0.563              & \textbf{414.048 $\pm$ 0.905}& -1.47\% \\
  \ac{fwd}    & \textbf{0.338 $\pm$ 0.017} & 0.398 $\pm$ 0.009             &  +17.75\% \\ \bottomrule       
  \end{tabular}
\end{table}

\subsection{Extended User Study}
For the first user study, following \cite{stein2023exposing}, we presented participants with pairs of real and generated images and asked them to select the realistic image. In this manner, we collected over 1k responses from 50 volunteers. In the second user study, we generated images from the Conceptual Captions~\citep{sharma2018conceptual} validation set and compared our metric with user alignments taken from \url{https://artificialanalysis.ai/text-to-image/arena?tab=Leaderboard}.
\begin{table}[!ht]
    \centering
    \caption{Comparison of metrics \ac{fid}, \ac{fd_dino} and \ac{fwd} with \ac{her}. Higher \ac{her} represents a higher prompt alignment percentage according to users. \ac{fwd} aligns better with \ac{her} than \ac{fid} and \ac{fd_dino}.}
    \label{tab:stable_diffusion_human}
    \begin{tabular}{ccccc}
    \toprule
         Generator & \ac{fid}$\downarrow$ & \ac{fd_dino}$\downarrow$ & \ac{fwd}$\downarrow$ & User Rating(\%)$\uparrow$ \\
    \midrule
         StableDiffusion-1.5 & \textbf{14.904} & \textbf{124.948} & 17.498 & 14  \\
         StableDiffusion-2.1 & 15.446 & 132.049	& 21.195 & 22\\
         StableDiffusion-3.0 (Medium) & 18.709 & 158.572	& 6.645 & 45\\
         StableDiffusion-3.5 (Large) & 17.907 & 157.798 & \textbf{4.979} & \textbf{61}\\
    \bottomrule
    \end{tabular}
\end{table}

\subsection{The fast wavelet and wavelet packet transforms}\label{sec:wpt}
This supplementary section summarizes key wavelet facts as a convenience for the reader.
See, for example, \citep{strang1996wavelets,mallat1999wavelet} or \citep{jensen2001ripples} for excellent detailed introductions to the topic.

The \acf{fwt} relies on convolution operations with filter pairs.
Figure~\ref{fig:fwt} illustrates the process. The forward or analysis transform
works with a low-pass $\mathbf{h}_\mathcal{L}$ and a high-pass filter $\mathbf{h}_\mathcal{H}$.
The analysis transform repeatedly convolves with both filters
\begin{align} \label{eq:fwt}
  \mathbf{x}_s *_1 \mathbf{h}_k = \mathbf{c}_{k, s+1}
\end{align}
with $*_1$ being the 1d-convolution operation, $k \in [\mathcal{L}, \mathcal{H}]$ and $s \in \mathbb{N}_0$, the set of natural numbers. While $\mathbf{x}_0$ is equal to
the original input signal $\mathbf{x}$, at higher scales, the \ac{fwt} uses the low-pass filtered result as input, i.e., $\mathbf{x}_s = \mathbf{c}_{\mathcal{L}, s}$ if $s > 0$. 
The dashed arrow in Figure~\ref{fig:fwt} indicates that we could continue to expand the \ac{fwt} tree here.
\begin{figure}
\centering
\includestandalone[scale=0.9]{./figures/supplementary/fwt}
\caption{Overview of the \acf{fwt} computation. $\mathbf{h}_\mathcal{L}$ denotes the analysis low-pass filter and $\mathbf{h}_\mathcal{H}$ the analysis high pass filter.  $\mathbf{f}_\mathcal{L}$ and $\mathbf{f}_\mathcal{H}$ 
the synthesis filter pair. $\downarrow_2$ denotes downsampling with a factor of two, $\uparrow_2$
means upsampling. The analysis transform relies on stride two convolutions.
The synthesis or inverse transform on the right works with stride two transposed convolutions.
$\mathbf{H}_{k}$ and $\mathbf{F}_{k}$ with $k \in [\mathcal{L}, \mathcal{H}]$ denote the corresponding convolution
operators.}
\label{fig:fwt}
\end{figure}

The \acf{wpt} additionally expands the high-frequency part of the tree.
\begin{figure}
\centering
\includestandalone[scale=0.9]{./figures/supplementary/packets_1d}  
\caption{Schematic drawing of the full \acf{wpt} in a single dimension.
Compared to Figure~\ref{fig:fwt}, the high-pass filtered side of the tree is expanded, too.}
\label{fig:wpt}
\end{figure}
A comparison of Figures~\ref{fig:fwt} and \ref{fig:wpt} illustrates this difference.
Whole expansion is not the only possible way to construct a wavelet packet tree. See \citep{jensen2001ripples} for a discussion of other options.
In both figures, capital letters denote convolution operators. These may be expressed as Toeplitz matrices \citep{strang1996wavelets}.
The matrix nature of these operators explains the capital boldface notation.
Coefficient subscripts record the path that leads to a particular coefficient.

We construct filter quadruples from the original filter pairs to process two-dimensional inputs. The process uses outer products \citep{vyas2018multiscale}:
\begin{align}\label{eq:2dfilters}
\mathbf{h}_{a} = \mathbf{h}_\mathcal{L}\mathbf{h}_\mathcal{L}^T,
\mathbf{h}_{h} = \mathbf{h}_\mathcal{L}\mathbf{h}_\mathcal{H}^T,
\mathbf{h}_{v} = \mathbf{h}_\mathcal{H}\mathbf{h}_\mathcal{L}^T,
\mathbf{h}_{d} = \mathbf{h}_\mathcal{H}\mathbf{h}_\mathcal{H}^T
\end{align}
With $a$ for approximation, $h$ for horizontal, $v$ for vertical, and $d$ for diagonal \citep{lee2019pywavelets}.
We can construct a \ac{wpt}-tree for images with these two-dimensional filters. 
\begin{figure}
\centering
\includestandalone[scale=0.9]{./figures/supplementary/packets_2d}  
\caption{Two dimensional \acf{wpt} computation overview. $\mathbf{X}$ and $\hat{\mathbf{X}}$ denote input image and reconstruction respectively. We compute the \acf{fwd} using the wavelet packet coefficients $\mathbf{p}$. The transform is invertible, the distance computation is therefore based on a lossless representation.}
\label{fig:wpt2d}
\end{figure}
Figure~\ref{fig:wpt2d} illustrates the computation of a full two-dimensional wavelet packet tree.
More formally, the process initially evaluates
\begin{align}
\mathbf{x}_0 * \mathbf{h}_j = \mathbf{c}_{j, 1}
\end{align}
with $\mathbf{x}_0$ equal to an input image $\mathbf{X}$, $j \in [a,h,v,d]$, and $*$ being the two-dimensional convolution. At higher scales, all resulting coefficients from previous scales serve as inputs. The four filters are repeatedly convolved with all outputs to build the full tree. The inverse transforms work analogously. We refer to the standard literature \citep{jensen2001ripples,strang1996wavelets} for an extended discussion.

Compared to the \ac{fwt}, the high-frequency half of the tree is subdivided into more bins, yielding a fine-grained view of the entire spectrum.
We always show analysis and synthesis transforms to stress that all wavelet transforms are lossless. Synthesis transforms reconstruct the original input based on the results from the analysis transform.

\subsection{Histogram matching - InceptionV3}
To understand the results in Table~\ref{tab:new_metrics} better, we present the histograms of InceptionV3 output labels for images in the datasets \ac{celebahq} and LSUN-Church in Figures~\ref{fig:celebahq_hist} and \ref{fig:lsun_church_hist}, respectively. In both figures, we compare the histograms of the generated images of \ac{ddpm} and StyleSwin. While StyleSwin generates better images than \ac{ddpm}, the class distribution of \ac{ddpm} is closer to the real images compared to StyleSwin on \ac{celebahq}. As a result, \ac{fid} is better for \ac{ddpm} in Table~\ref{tab:new_metrics}. For LSUN-Church, the distributions are more similar and \ac{fid} correctly estimates that StyleSwin generates better images than \ac{ddpm}. In contrast to \ac{fid}, \ac{fwd} is not fooled by the class distribution and provides a consistent ranking for \ac{ddpm} and StyleSwin on both datasets, as reported in Table~\ref{tab:new_metrics}. 

\begin{figure}[!h]
  \centering
  \begin{subfigure}[b]{0.4\textwidth}
    \includestandalone[width=\textwidth]{./figures/histograms/celebahq_hist}
    \caption{CelebAHQ}
    \label{fig:celebahq_hist}
  \end{subfigure}
  \hfill
  \begin{subfigure}[b]{0.4\textwidth}
    \includestandalone[width=\textwidth]{./figures/histograms/lsun_church_hist}
    \caption{LSUN-Church}
    \label{fig:lsun_church_hist}
  \end{subfigure}
  \label{fig:hist_celeb_lsun}
  \caption{Histograms of predicted top-1 classes by the InceptionV3 network. 
  }
\end{figure}


\subsection{Compute details}
While the proposed evaluation metric \ac{fwd} is very efficient, some of the generative models are expensive. We used 16 nodes with 4 Nvidia A100 GPUs to generate the samples in Table~\ref{tab:new_metrics}.

\subsection{\ac{agriculture}}
 \ac{agriculture} contains 3600 images with 7 classes and the task requires detecting nutrient deficiency in winter wheat and winter rye, such as nitrogen, phosphorous, and potassium deficiencies. The images were captured over the 2019 growth period at the long-term fertilizer experiment (LTFE) Dikopshof near Bonn and were annotated with seven types of fertilizer treatments. We preprocessed the dataset by splitting the 1000x1000 resolution image into 256x256 crops. This resulted in 57600 images overall.
We trained \ac{projgan} and \ac{ddgan} on this preprocessed dataset.
  

\subsection{Sentinel Dataset}
The Sentinel dataset consists of 180,662 triplets of Synthetic Aperture Radar (SAR) image patches collected from Sentinel-1 and Sentinel-2 missions. From these, we only use the images from the ROIs\_2017\_Winter subset, which contain 31,825 images. We train \ac*{projgan} and \ac*{ddgan} on this subset. The original images are stored in the "tif" format and conversion to "jpg" is made using the official codebase provided by \cite{schmitt2019sen12ms}.

\subsection{Additional metrics}
\label{ref:extend_metrics}
Here, we present the comparison with additional metrics such as \ac{fid}$_\infty$~\citep{chong20fidbash}, \ac{is}~\citep{salimans2016improved}, \ac{is}$_\infty$~\citep{chong20fidbash}, Clean-FID~\citep{parmar2022aliased}, and \ac{kid}~\citep{binkowski18kid}. Table~\ref{tab:extend_metrics} extends the results from Table~\ref{tab:inet_bias_values}. The results show that all the stated metrics suffer from domain bias, as they share the same ImageNet pretrained Inception-V3 backbone.

\begin{table}[!ht]
    \centering
    \caption{Extended comparison of metrics to detect domain bias. All the metrics which share the pretrained InceptionV3 backbone suffer from domain bias, whereas \ac{fwd} is domain agnostic.}
    \label{tab:extend_metrics}
    \resizebox{\textwidth}{!}{
    \begin{tabular}{cccccccccc}
    \toprule
        ~ & ~ & FID$_\infty$$\downarrow$ & KID$\downarrow$ & Clean-FID$\downarrow$ & \ac{is}$\uparrow$ & IS$_\infty$$\uparrow$ & \ac{fid}$\downarrow$ & \ac{fd_dino}$\downarrow$ & FWD$\downarrow$ \\ 
    \midrule
        \multirow{2}{*}{\ac{celebahq}} & Proj. FastGAN & \textbf{6.222} & \textbf{0.0020} & \textbf{6.729} & \textbf{2.925} & \textbf{2.545} & \textbf{6.358} & 685.889 & 1.388 \\ 
        ~ & DDGAN & 6.961 & 0.0034 & 7.156 & 2.669 & 2.315 & 7.641 & \textbf{199.761} & \textbf{0.408} \\ 
    \midrule
        \multirow{2}{*}{FFHQ} & Proj. FastGAN & \textbf{4.048} & \textbf{0.0006} & \textbf{4.206} & \textbf{5.358} & \textbf{3.732} & \textbf{4.106} & 593.124 & 0.651 \\ 
        ~ & StyleGAN2 & 4.782 & 0.0011 & 4.597 & 5.307 & 3.714 & 4.282 & \textbf{420.273} & \textbf{0.312} \\ 
    \midrule
        \multirow{2}{*}{\ac{agriculture}} & Proj. FastGAN & \textbf{5.141} & \textbf{0.0032} & \textbf{5.597} & \textbf{2.461} & \textbf{2.142} & \textbf{4.675} & \textbf{171.625} & 1.442 \\ 
        ~ & DDGAN & 25.872 & 0.025 & 26.427 & 2.332 & 2.105 & 26.233 & 232.884 & \textbf{1.357} \\
    \midrule
        \multirow{2}{*}{Sentinel} & Proj. FastGAN & \textbf{5.216} & \textbf{0.0030} & \textbf{9.087} & \textbf{3.846} & 3.257 & \textbf{8.96} & 424.898 & 0.755 \\ 
        ~ & DDGAN & 26.154 & 0.0248 & 23.358 & 3.647 & \textbf{3.329} & 23.615 & \textbf{404.700} & \textbf{0.115} \\
        \bottomrule
    \end{tabular}}
\end{table}

In addition, Table~\ref{tab:swd_metric} presents the results of \ac{swd}~\citep{karras2018progGAN} and \ac{fwd} for generated \ac{celebahq} images from \ac{projgan} and \ac{ddgan} where we compute each metric five times independently. While \ac{swd} is robust to the domain bias, the randomized projections lead to a very high standard deviation, making this metric unreliable in practice.  Our proposed metric \ac{fwd} is deterministic and provides in all runs the same result.
\begin{table}[!ht]
    \centering
    \caption{Reproducibility of FWD and SWD. We report minimum and mean $\pm$ standard deviation in brackets across 5 independent runs.}
    \label{tab:swd_metric}
    \begin{tabular}{cccc}
        \toprule
        Dataset & Generator & \ac{fwd}$\downarrow$ & \ac{swd}$\downarrow$ \\
        \midrule
         \multirow{2}{*}{\ac{celebahq}}& \ac{projgan} & 1.388 (1.388 $\pm$ 0.00) &  169.694 (175.292 $\pm$ 3.54) \\
          & \ac{ddgan} & \textbf{0.408 (0.408  $\pm$ 0.00)} & \textbf{99.198 (108.553$\pm$6.81)} \\
        \bottomrule
    \end{tabular}
\end{table}

\subsection{FID pretrained with DNDD}
As discussed in the results section, fine-tuning the InceptionV3 backbone with \ac{agriculture} does not solve the domain bias problem. Since the dataset consists of only 3600 images, the InceptionV3 network fails to learn representative features to compute \ac{fid}. Table~\ref{tab:fid_dndd} shows that fine-tuned \ac{fid} still prefers \ac{projgan}.
\begin{table}[!ht]
  \centering
  \caption{Comparison of \ac{fid} (ImageNet), \ac{fid} (DNDD) and \ac{fwd} on \ac{agriculture}. After fine-tuning InceptionV3 on \ac{agriculture}, \ac{fid} (DNDD) still prefers \ac{projgan} whereas \ac{fwd} ranks \ac{ddgan} better.}
  \label{tab:fid_dndd}
  \begin{tabular}{ccccc}
    \toprule
    Dataset & Generator & \acs{fid} (ImageNet)$\downarrow$ & \acs{fid} (DNDD)$\downarrow$ & \acs{fwd}$\downarrow$ \\ \midrule
    \multirow{2}{*}{\acs{agriculture}} & \acs{projgan} & \textbf{4.675}  & \textbf{20.937} & 1.442 \\
                                       & \acs{ddgan}   & 26.233 & 52.521 & \textbf{1.357} \\
  \bottomrule
  \end{tabular}
\end{table}

\section{Extended Discussion of Related Work}
\label{sec:extended_related_work}
\subsection{Spectral Methods}
\label{spectral_methods}
Prior work found neural networks are spectrally biased \citep{rahaman2019spectral} and many architectures favor low-frequency content \citep{durall2020watch,gal2021swagan,wolter2022wavelet, zhang2022styleswin}. Related articles rely on the Fourier or Wavelet transform to understand frequency bias. Wavelet transforms, as pioneered by \cite{Mallat89Wavelet} and \cite{Daubechies92Ten}, have a solid track record in signal processing. The \acf{fwt} and the closely related \acf{wpt} are starting to appear more frequently in the deep learning literature. Applications include \ac{cnn} augmentation \citep{williams2018wavelet}, style transfer \citep{yoo2019photorealistic}, image denoising \citep{liu2020wavelet,saragadam2023wire}, image coloring \citep{li2022wavelet}, face aging \citep{liu2019attribute}, video enhancement \citep{wang2020multi}, face super-resolution \citep{huang2017wavelet}, and generative machine learning \citep{gal2021swagan,guth2022wavelet,zhang2022styleswin,Phung23WaveDiff}. \cite{hernandez2019human} use the Fourier transform to measure the quality of human motion forecasting. \cite{zhang2022styleswin} use a \ac{fwt} to remove artifacts from generated images. \cite{Phung23WaveDiff} focuse on the \ac{fwt} to increase the inference speed of diffusion models. 
This work proposes to use the \acf{wpt} as an interpretable metric for generators.

\subsection{Generative Architectures}
\label{generative_vision}


Prior work mainly falls into the three \ac{gan}, Diffusion, and \ac{vae} architecture groups. The StyleGAN architecture family \citep{karras2019style,karras2020analyzing,karras2021alias} is among the pioneering architectures in generative vision. \ac{gan}s allow rapid generation of high-quality images but suffer from training instability and poor mode coverage \citep{salimans2016improved}. \cite{Sauer2021NEURIPS} proposed the \acf{projgan} architecture, which stabilizes and improves training convergence by introducing ImageNet pre-trained weights into the discriminator. The upgraded discriminator pushes the output distribution towards ImageNet.
\ac{vae} models \citep{kingma14auto,van2017neural}, on the other hand,  enable the generation of diverse image sets, but are unable to produce high-quality images. 

Diffusion models \citep{sohl2015deep,ho20ddpm,peebles2023scalable} have emerged as a very promising alternative and produce high-quality images \citep{ho20ddpm,dhariwal2021diffusion} in an autoregressive style. \acp{ddpm}, for example, are Markovian processes that learn to gradually separate added noise from data during training. During inference, images are generated from Gaussian noise via a reverse process that requires iterating through all steps to generate an image. \cite{song2021ddim} reduced the number of sampling steps by introducing \ac{ddim}, which relies on a deterministic non-Markovian sampling process.
Furthermore, \cite{nichol2021improvdiff} proposed the use of strided sampling, to reduce the sampling timesteps and also provide a performance improvement by using cosine instead of linear sampling. Moreover, \cite{nichol2021improvdiff} adopt a weighted variational lower bound to supplement the \ac{mse} loss. 
In an attempt to solve the generative learning trilemma (image quality, diversity and fast sampling), \cite{xiao2022tackling} proposed \acf{ddgan}. The paper parameterizes a conditional GAN for the reverse diffusion process and demonstrates faster generation speed.

\section{Additional Samples}
\subsection{Stable Diffusion}
\begin{figure}
    \centering
    \includegraphics[width=\textwidth]{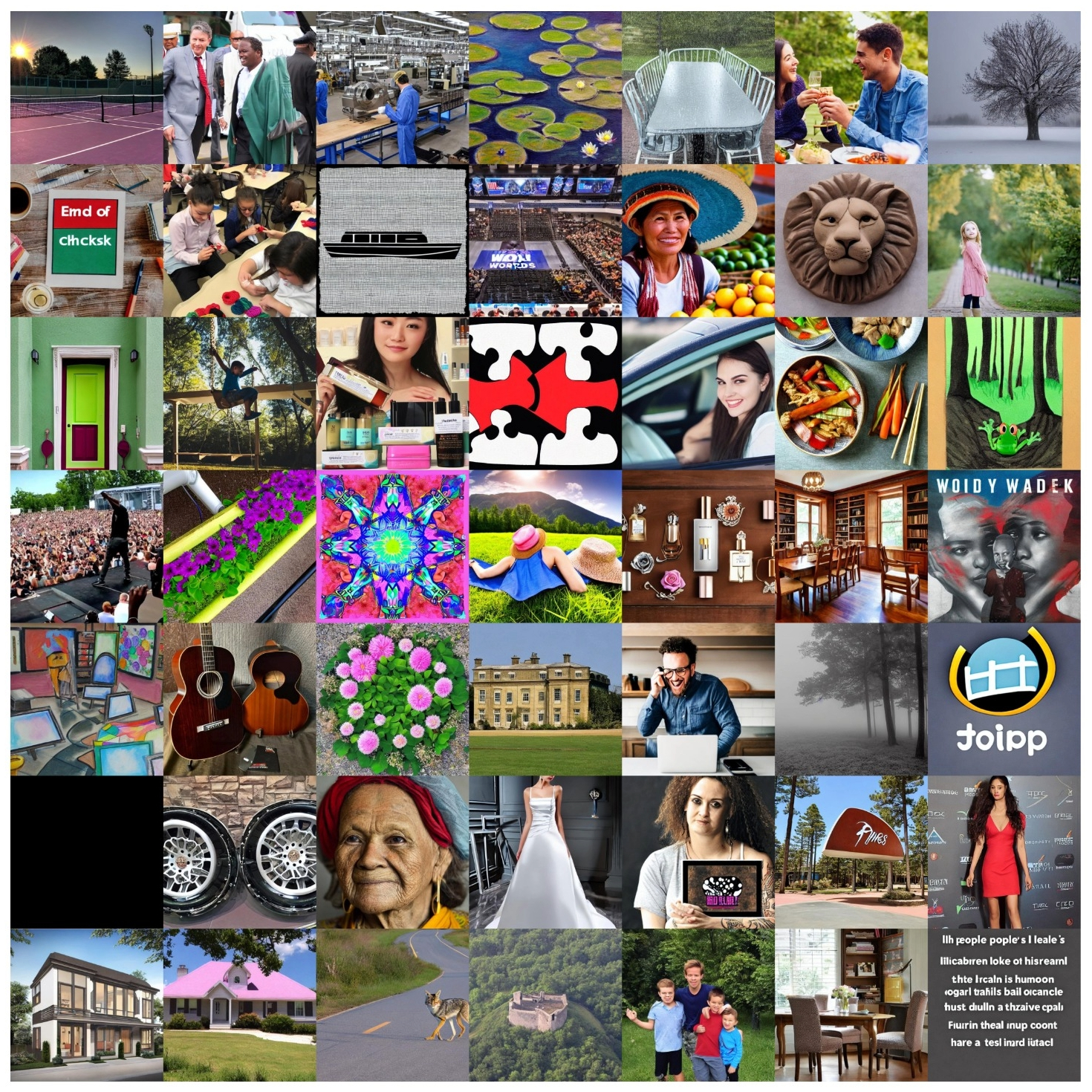}
    \caption{Samples from StableDiffusion v1.5 generated with prompts from the Conceptual Caption validation set.}
    \label{fig:sd_1_5}
\end{figure}

\begin{figure}
    \centering
    \includegraphics[width=\textwidth]{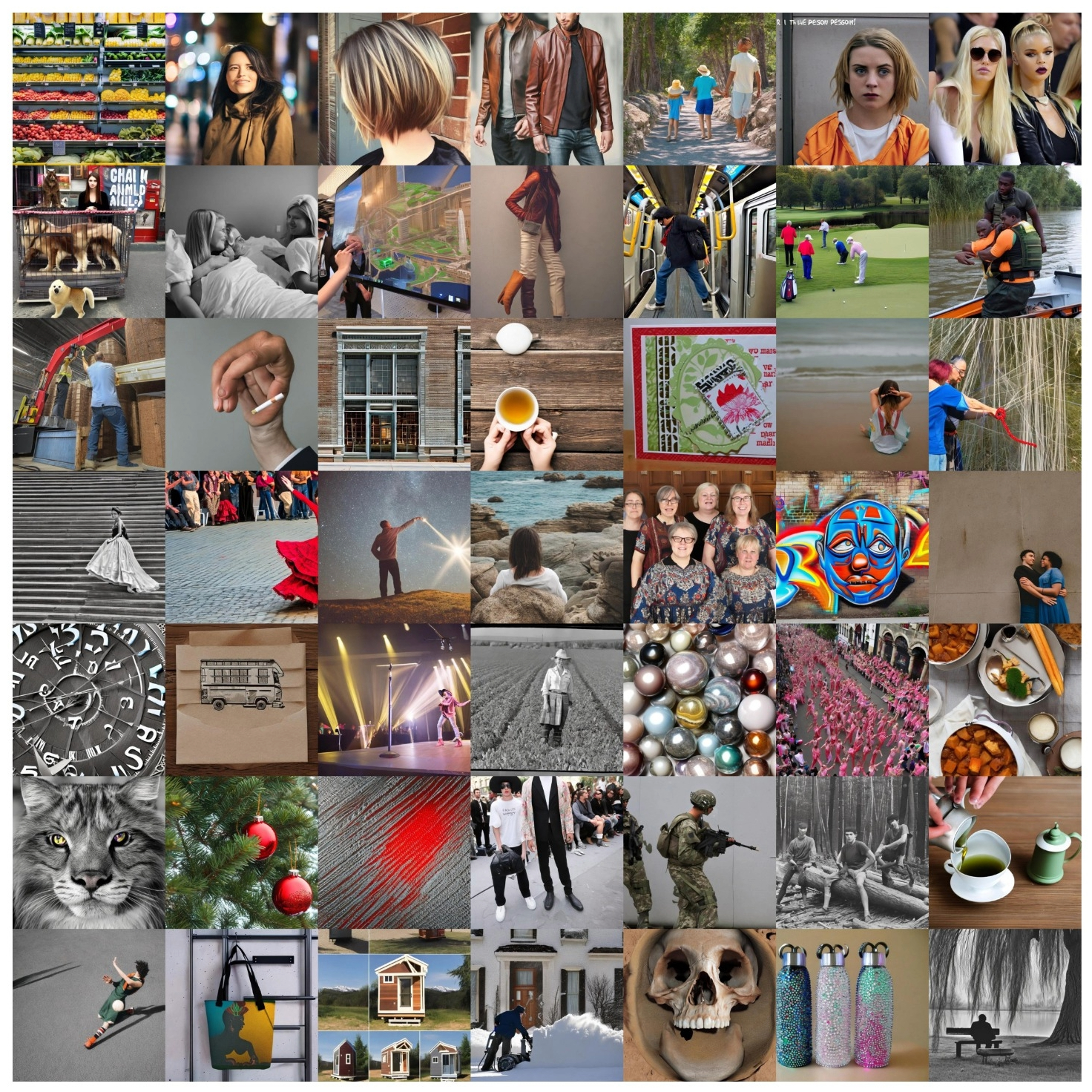}
    \caption{Samples from StableDiffusion v2.1 generated with prompts from the Conceptual Caption validation set.}
    \label{fig:sd_2_1}
\end{figure}

\begin{figure}
    \centering
    \includegraphics[width=\textwidth]{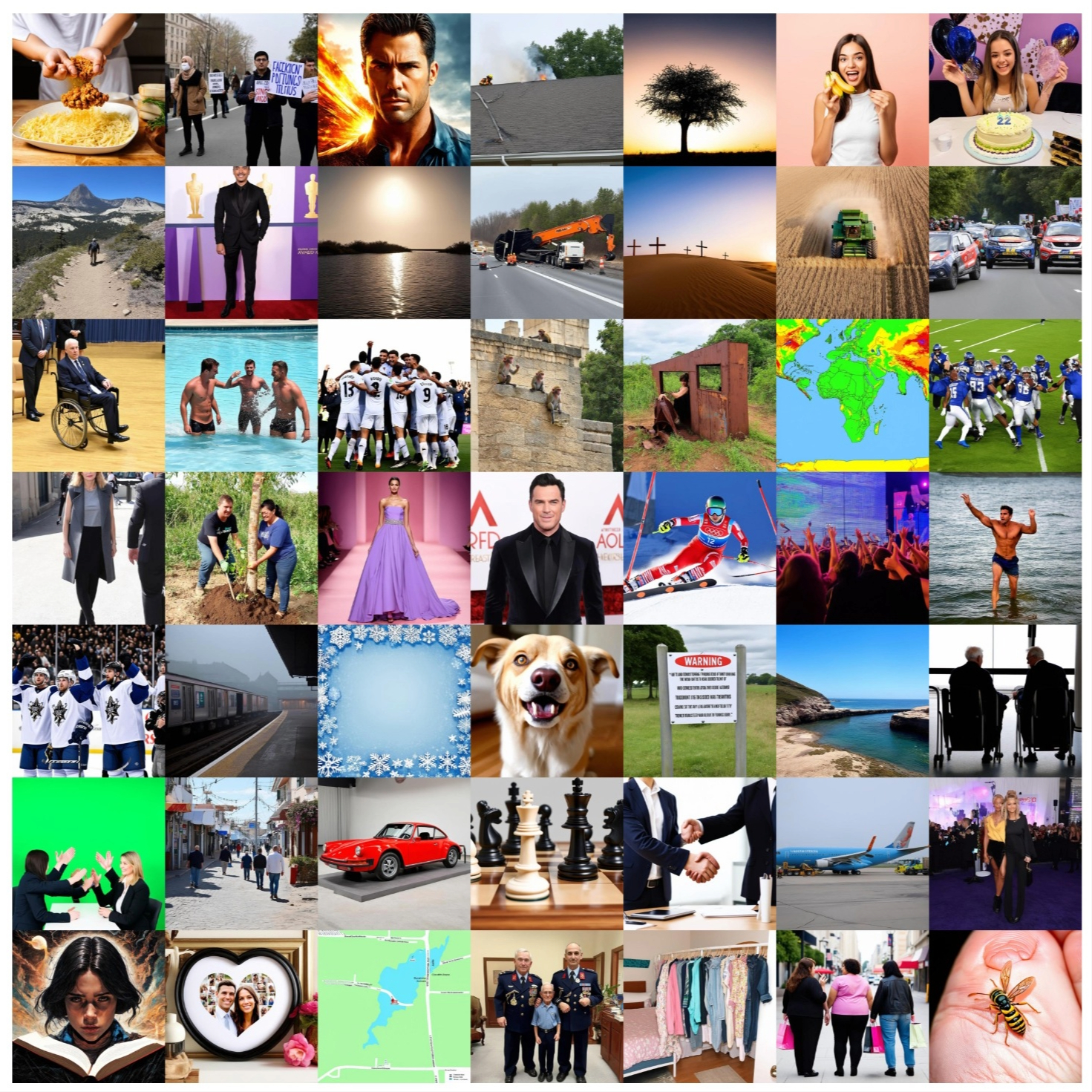}
    \caption{Samples from StableDiffusion v3.0 (Medium) generated with prompts from the Conceptual Caption validation set.}
    \label{fig:sd_3_0}
\end{figure}

\begin{figure}
    \centering
    \includegraphics[width=\textwidth]{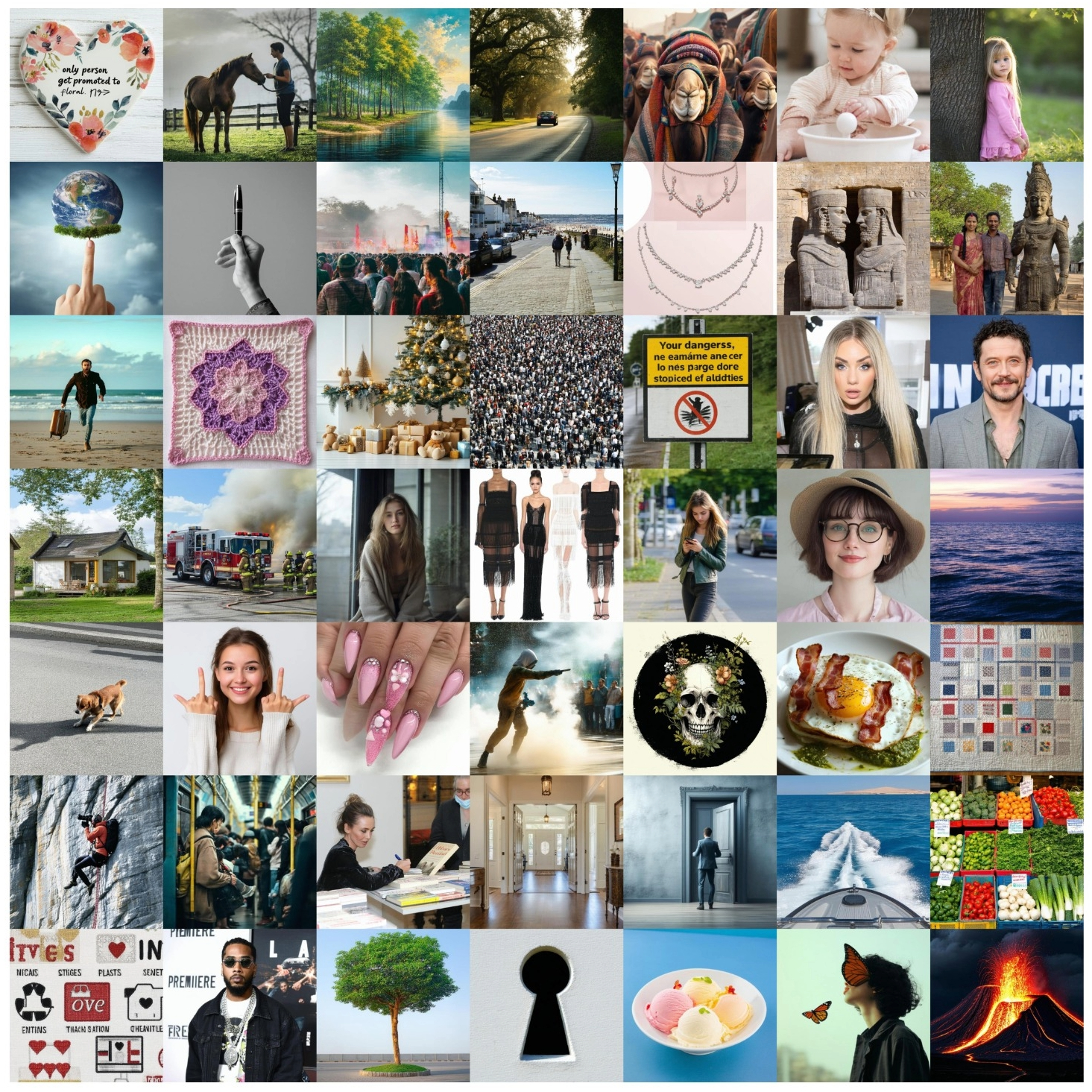}
    \caption{Samples from StableDiffusion v3.5 (Large) generated with prompts from the Conceptual Caption validation set.}
    \label{fig:sd_3_5}
\end{figure}

In this section, we provide the generated samples from StableDiffusion models used for user study when evaluated on the Conceptual Captions dataset~\citep{sharma2018conceptual}. In particular, we use versions 1.5, 2.1, 3.0 (Medium), and 3.5 (Large), and Figures~\ref{fig:sd_1_5},~\ref{fig:sd_2_1},~\ref{fig:sd_3_0}, and~\ref{fig:sd_3_5} represent the samples from these models respectively. 
\label{sec:sd_samples}

\subsection{DDGAN and Proj.FastGAN}
\begin{figure}
  \includegraphics[width=\textwidth]{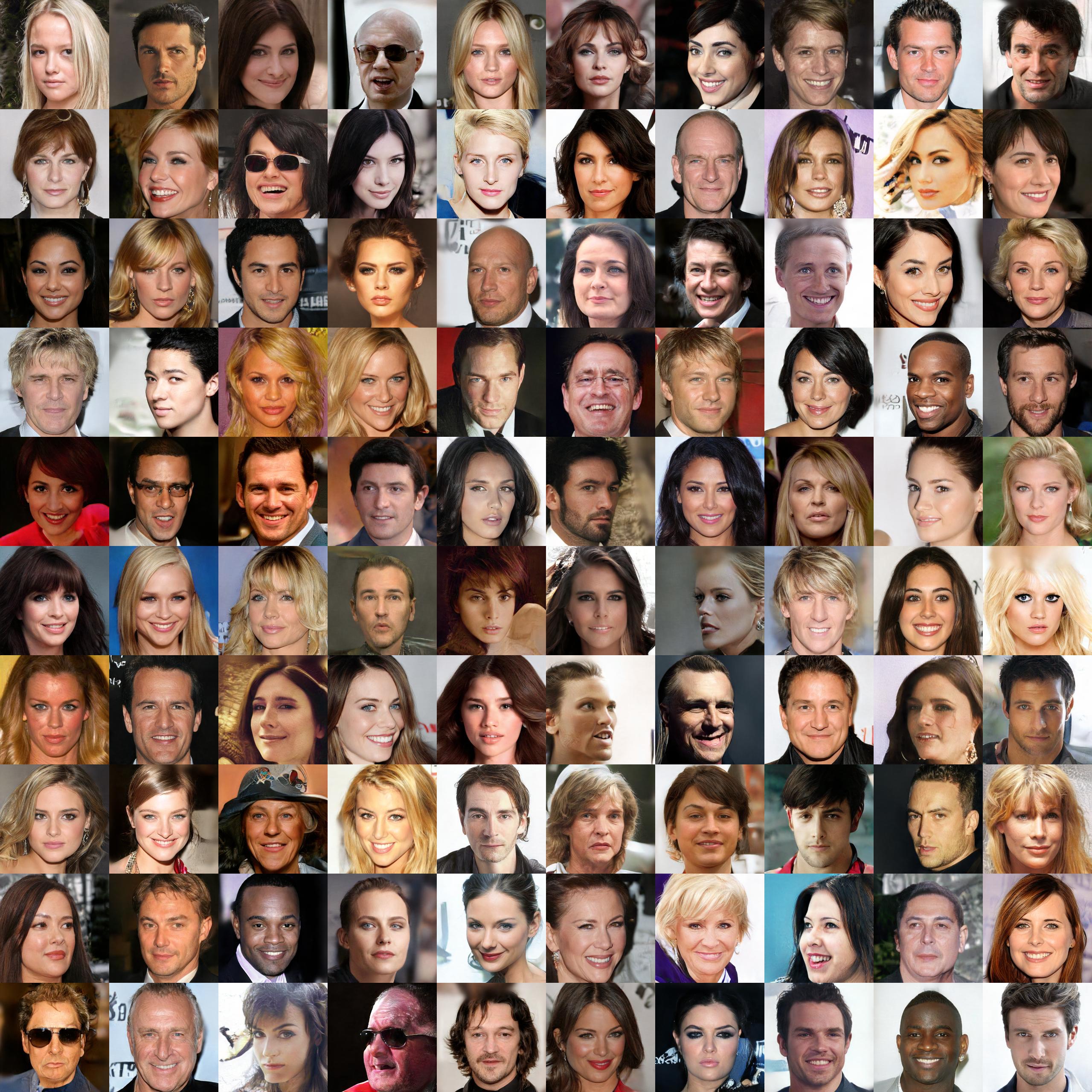}
  \caption{Samples from \ac{ddgan} trained on the \ac{celebahq} dataset.}
  \label{fig:ddgan_celebahq_full}
\end{figure}

\begin{figure}
  \includegraphics[width=\textwidth]{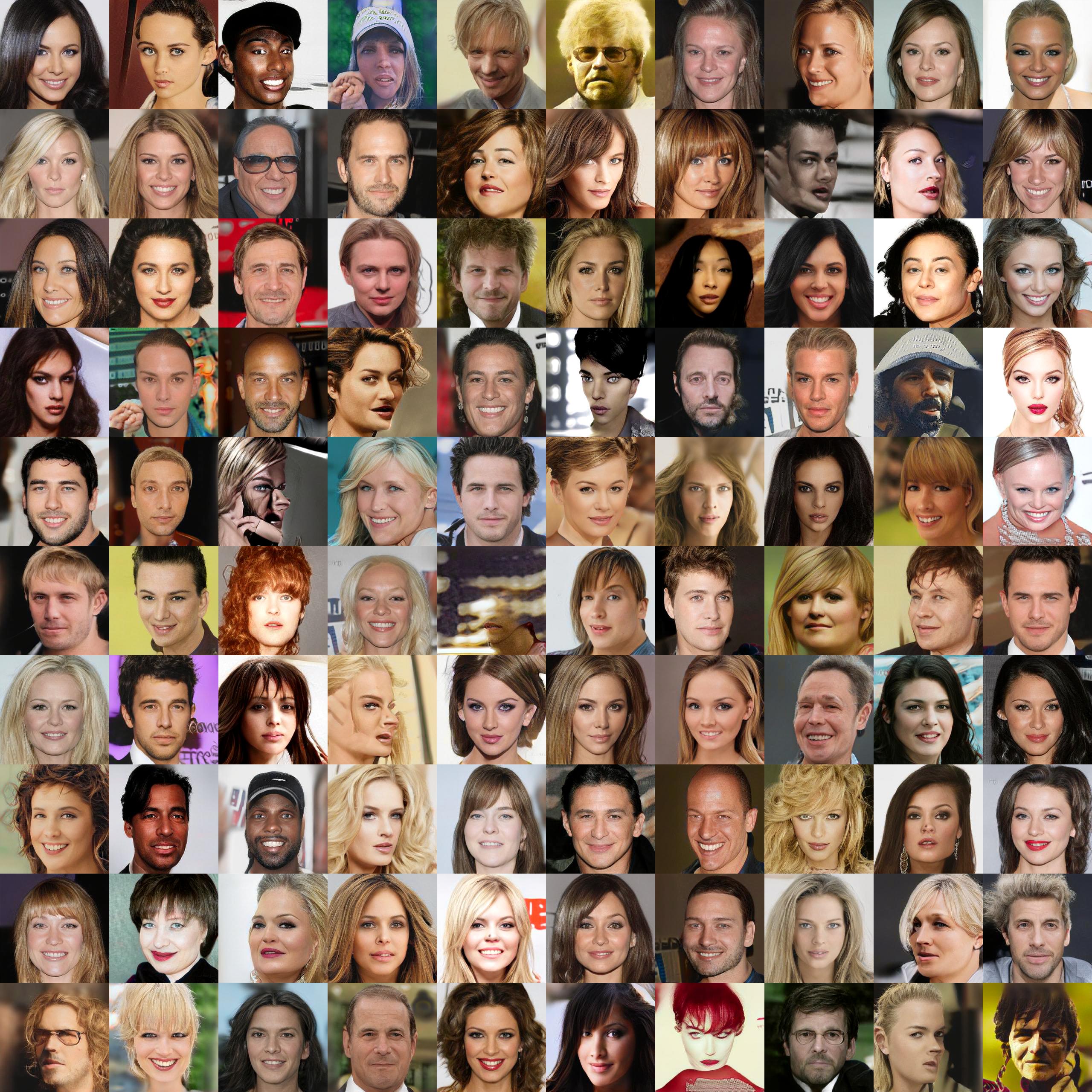}
  \caption{Samples from \ac{projgan} trained on the \ac{celebahq} dataset.}
  \label{fig:pggan_celebahq_full}
\end{figure}

\begin{figure}
  \includegraphics[width=\textwidth]{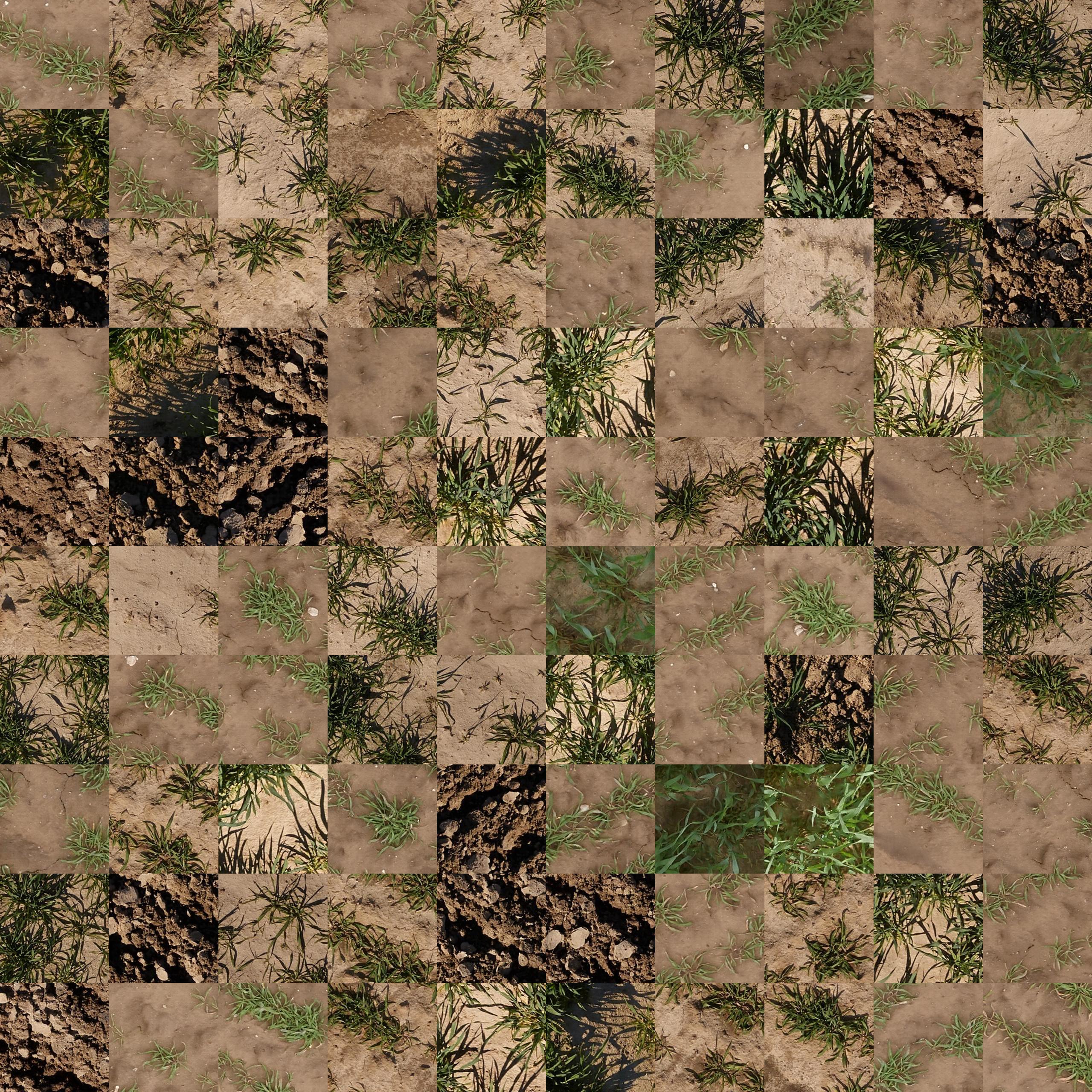}
  \caption{Samples from \ac{ddgan} trained on the \ac{agriculture}.}
  \label{fig:ddgan_dndd_full}
\end{figure}

\begin{figure}
  \includegraphics[width=\textwidth]{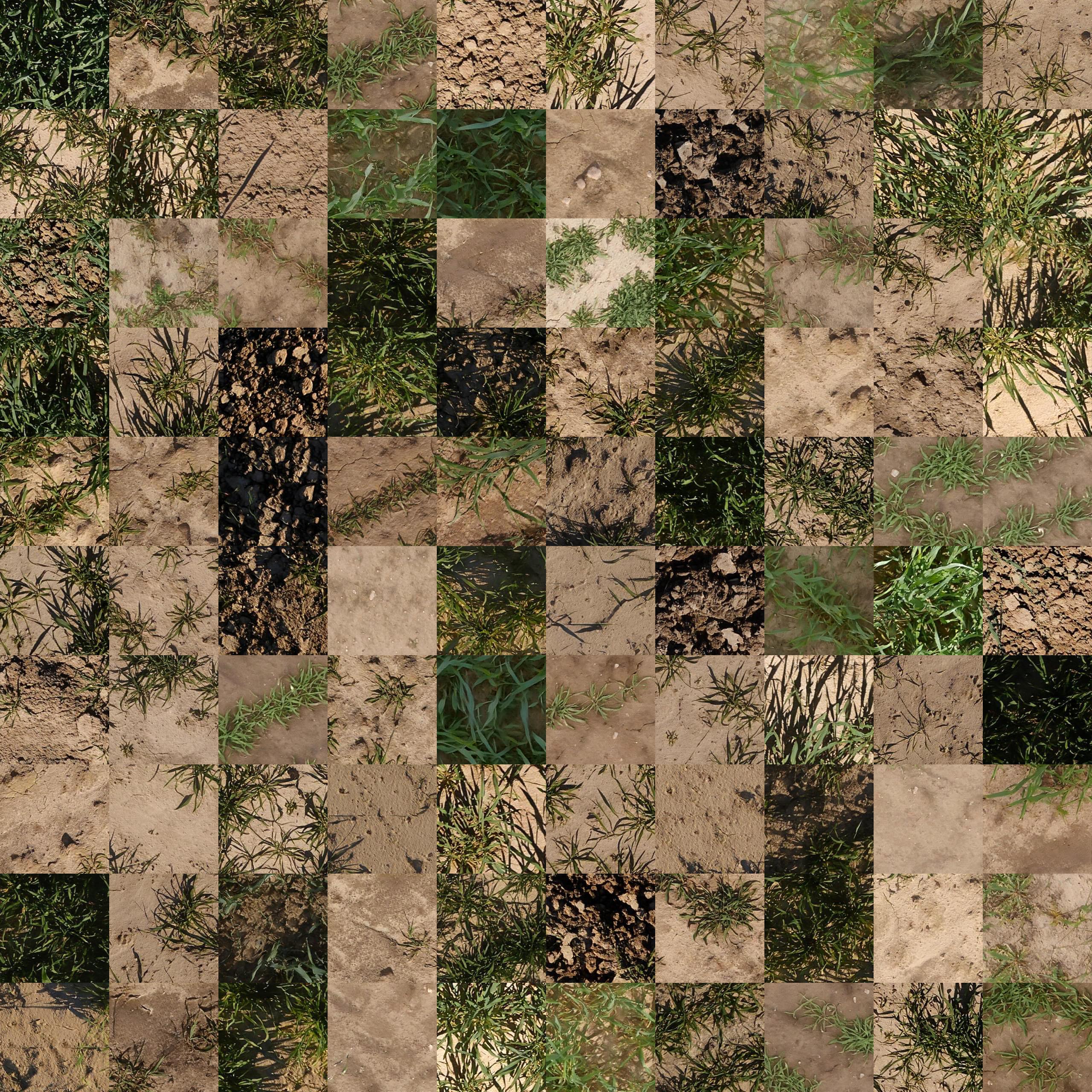}
  \caption{Samples from \ac{projgan} trained on the \ac{agriculture}.}
  \label{fig:pggan_dndd_full}
\end{figure}

\begin{figure}
  \includegraphics[width=\textwidth]{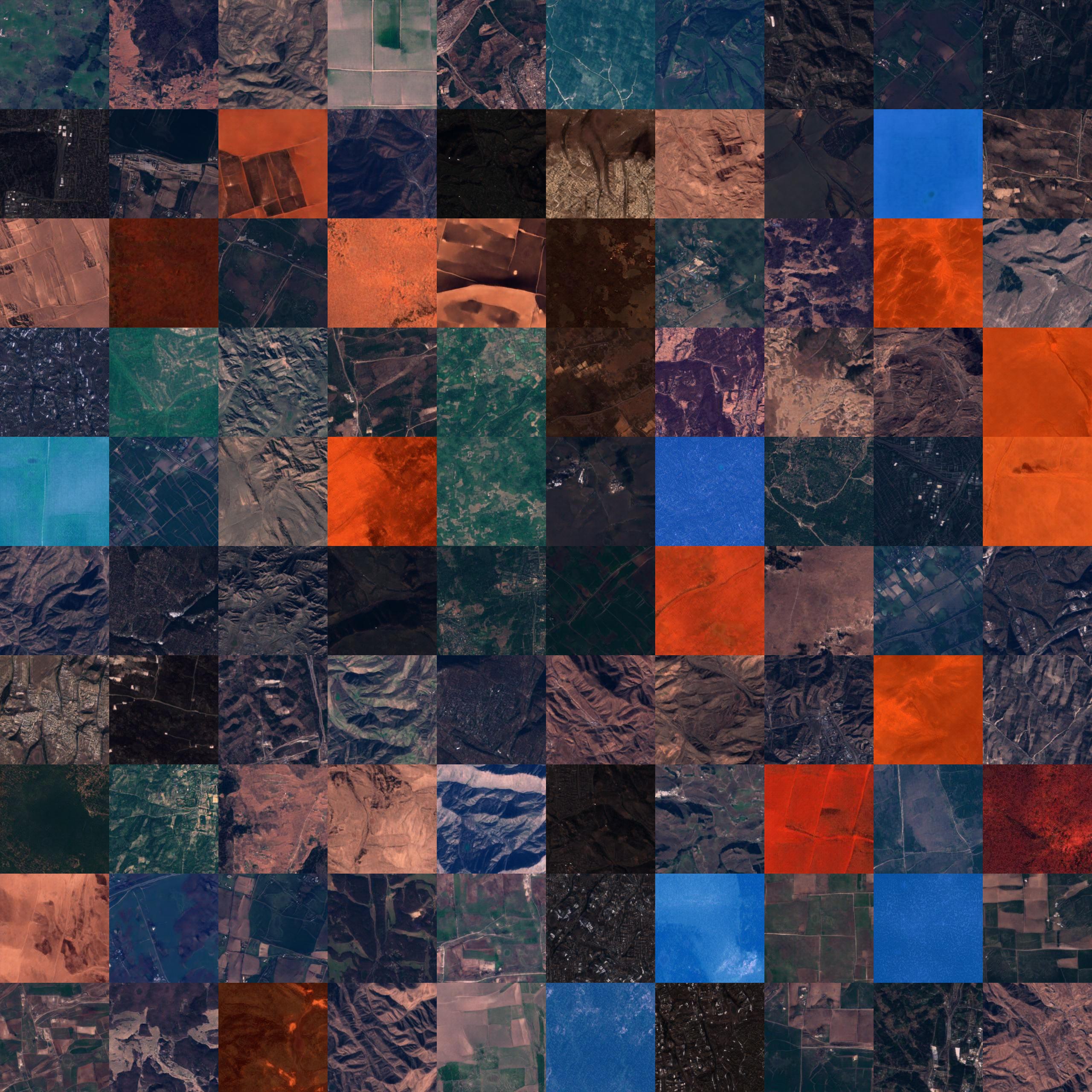}
  \caption{Samples from \ac{ddgan} trained on the Sentinel dataset.}
  \label{fig:ddgan_sent_full}
\end{figure}

\begin{figure}
  \includegraphics[width=\textwidth]{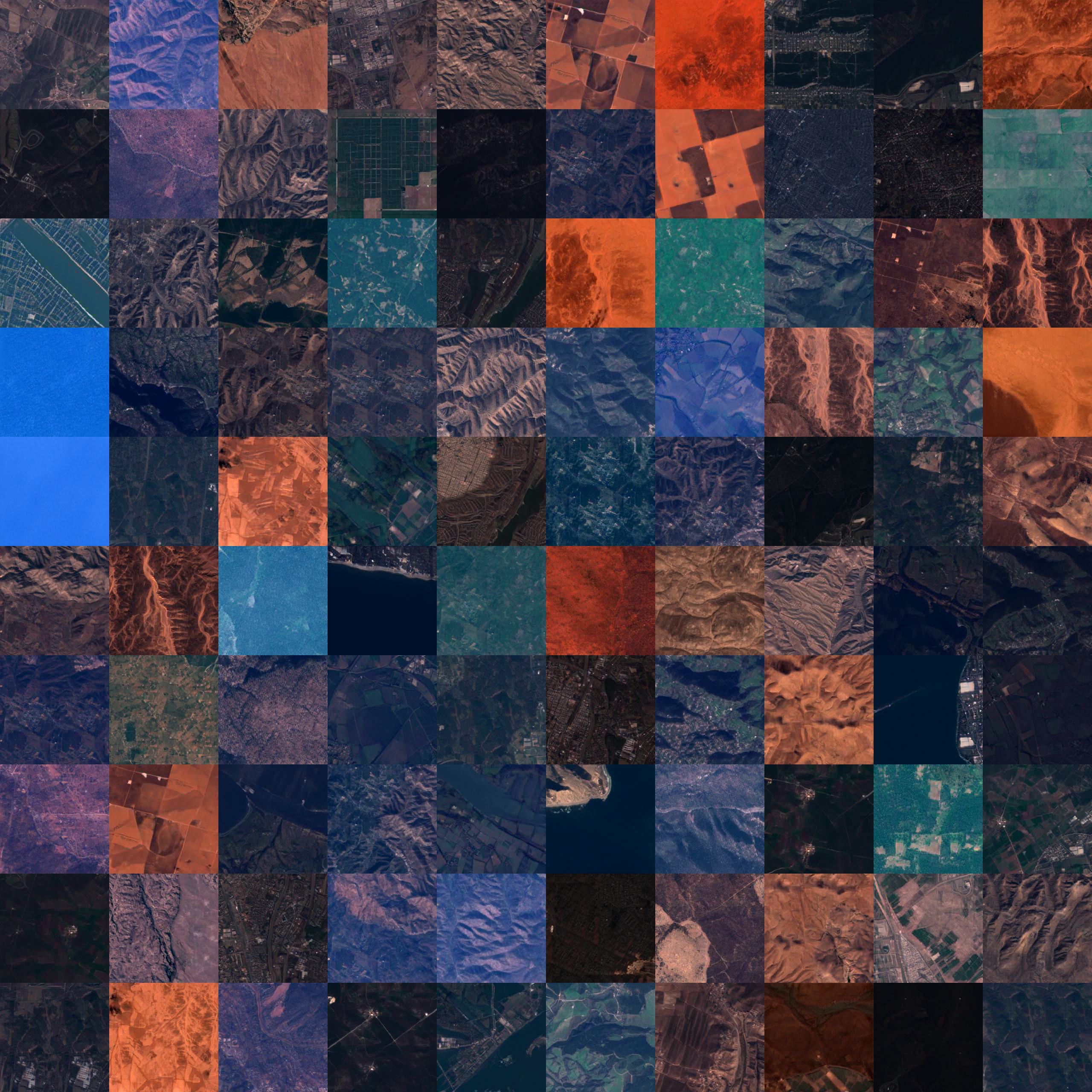}
  \caption{Samples from \ac{projgan} trained on the Sentinel dataset.}
  \label{fig:pggan_sent_full}
\end{figure}

Here we present the additional samples generated from \ac{ddgan} and \ac{projgan} trained on \ac{celebahq}, DNDD and Sentinel datasets individually. Figures~\ref{fig:ddgan_celebahq_full}, \ref{fig:pggan_celebahq_full} represent CelebAHQ samples from DDGAN and \ac{projgan} respectively. Similarly, Figures~\ref{fig:ddgan_dndd_full}, \ref{fig:pggan_dndd_full} and Figures~\ref{fig:ddgan_sent_full}, \ref{fig:pggan_sent_full} depict samples from DNDD and Sentinel datasets, respectively.

\end{document}

%% file: math_commands.tex
\usepackage{amsmath,amsfonts,bm}

\def\eqref#1{equation~\ref{#1}}

\def\1{\bm{1}}

\DeclareMathAlphabet{\mathsfit}{\encodingdefault}{\sfdefault}{m}{sl}
\SetMathAlphabet{\mathsfit}{bold}{\encodingdefault}{\sfdefault}{bx}{n}



%% file: figures/histograms/celeba_hist.tex
\begin{tikzpicture}

\definecolor{color0}{rgb}{0.12156862745098,0.466666666666667,0.705882352941177}
\definecolor{color1}{rgb}{1,0.498039215686275,0.0549019607843137}
\definecolor{color2}{rgb}{0.172549019607843,0.627450980392157,0.172549019607843}

\begin{axis}[
legend cell align={left},
legend style={fill opacity=0.8, draw opacity=1, text opacity=1, draw=white!80!black},
tick align=outside,
tick pos=left,
x grid style={white!69.0196078431373!black},
xmin=-0.832, xmax=15.272,
xtick style={color=black},
xtick={0.44,1.44,2.44,3.44,4.44,5.44,6.44,7.44,8.44,9.44,10.44,11.44,12.44,13.44,14.44},
xticklabel style={rotate=90.0,anchor=east},
xticklabels={
  wig,
  bikini,
  suit,
  bow tie,
  lipstick,
  brassiere,
  seat belt,
  sunscreen,
  fur coat,
  neck brace,
  lab coat,
  Band Aid,
  hair spray,
  stethoscope,
  sunglasses
},
y grid style={white!69.0196078431373!black},
ymin=0, ymax=11623.5,
ytick style={color=black}
]
\draw[draw=none,fill=color0] (axis cs:-0.1,0) rectangle (axis cs:0.1,6035);
\addlegendimage{ybar,ybar legend,draw=none,fill=color0}
\addlegendentry{CelebAHQ}

\draw[draw=none,fill=color0] (axis cs:0.9,0) rectangle (axis cs:1.1,2853);
\draw[draw=none,fill=color0] (axis cs:1.9,0) rectangle (axis cs:2.1,2822);
\draw[draw=none,fill=color0] (axis cs:2.9,0) rectangle (axis cs:3.1,1860);
\draw[draw=none,fill=color0] (axis cs:3.9,0) rectangle (axis cs:4.1,1662);
\draw[draw=none,fill=color0] (axis cs:4.9,0) rectangle (axis cs:5.1,1558);
\draw[draw=none,fill=color0] (axis cs:5.9,0) rectangle (axis cs:6.1,1109);
\draw[draw=none,fill=color0] (axis cs:6.9,0) rectangle (axis cs:7.1,1047);
\draw[draw=none,fill=color0] (axis cs:7.9,0) rectangle (axis cs:8.1,911);
\draw[draw=none,fill=color0] (axis cs:8.9,0) rectangle (axis cs:9.1,659);
\draw[draw=none,fill=color0] (axis cs:9.9,0) rectangle (axis cs:10.1,615);
\draw[draw=none,fill=color0] (axis cs:10.9,0) rectangle (axis cs:11.1,512);
\draw[draw=none,fill=color0] (axis cs:11.9,0) rectangle (axis cs:12.1,466);
\draw[draw=none,fill=color0] (axis cs:12.9,0) rectangle (axis cs:13.1,416);
\draw[draw=none,fill=color0] (axis cs:13.9,0) rectangle (axis cs:14.1,375);
\draw[draw=none,fill=color1] (axis cs:0.12,0) rectangle (axis cs:0.32,11070);
\addlegendimage{ybar,ybar legend,draw=none,fill=color1}
\addlegendentry{DDGAN}

\draw[draw=none,fill=color1] (axis cs:1.12,0) rectangle (axis cs:1.32,5895);
\draw[draw=none,fill=color1] (axis cs:2.12,0) rectangle (axis cs:2.32,8315);
\draw[draw=none,fill=color1] (axis cs:3.12,0) rectangle (axis cs:3.32,1030);
\draw[draw=none,fill=color1] (axis cs:4.12,0) rectangle (axis cs:4.32,305);
\draw[draw=none,fill=color1] (axis cs:5.12,0) rectangle (axis cs:5.32,7208);
\draw[draw=none,fill=color1] (axis cs:6.12,0) rectangle (axis cs:6.32,658);
\draw[draw=none,fill=color1] (axis cs:7.12,0) rectangle (axis cs:7.32,237);
\draw[draw=none,fill=color1] (axis cs:8.12,0) rectangle (axis cs:8.32,1103);
\draw[draw=none,fill=color1] (axis cs:9.12,0) rectangle (axis cs:9.32,2033);
\draw[draw=none,fill=color1] (axis cs:10.12,0) rectangle (axis cs:10.32,1078);
\draw[draw=none,fill=color1] (axis cs:11.12,0) rectangle (axis cs:11.32,110);
\draw[draw=none,fill=color1] (axis cs:12.12,0) rectangle (axis cs:12.32,287);
\draw[draw=none,fill=color1] (axis cs:13.12,0) rectangle (axis cs:13.32,348);
\draw[draw=none,fill=color1] (axis cs:14.12,0) rectangle (axis cs:14.32,302);
\draw[draw=none,fill=color2] (axis cs:0.34,0) rectangle (axis cs:0.54,7214);
\addlegendimage{ybar,ybar legend,draw=none,fill=color2}
\addlegendentry{Proj. FastGAN}

\draw[draw=none,fill=color2] (axis cs:1.34,0) rectangle (axis cs:1.54,2835);
\draw[draw=none,fill=color2] (axis cs:2.34,0) rectangle (axis cs:2.54,4626);
\draw[draw=none,fill=color2] (axis cs:3.34,0) rectangle (axis cs:3.54,681);
\draw[draw=none,fill=color2] (axis cs:4.34,0) rectangle (axis cs:4.54,361);
\draw[draw=none,fill=color2] (axis cs:5.34,0) rectangle (axis cs:5.54,3764);
\draw[draw=none,fill=color2] (axis cs:6.34,0) rectangle (axis cs:6.54,339);
\draw[draw=none,fill=color2] (axis cs:7.34,0) rectangle (axis cs:7.54,93);
\draw[draw=none,fill=color2] (axis cs:8.34,0) rectangle (axis cs:8.54,358);
\draw[draw=none,fill=color2] (axis cs:9.34,0) rectangle (axis cs:9.54,945);
\draw[draw=none,fill=color2] (axis cs:10.34,0) rectangle (axis cs:10.54,540);
\draw[draw=none,fill=color2] (axis cs:11.34,0) rectangle (axis cs:11.54,78);
\draw[draw=none,fill=color2] (axis cs:12.34,0) rectangle (axis cs:12.54,165);
\draw[draw=none,fill=color2] (axis cs:13.34,0) rectangle (axis cs:13.54,265);
\draw[draw=none,fill=color2] (axis cs:14.34,0) rectangle (axis cs:14.54,351);
\end{axis}

\end{tikzpicture}

%% file: figures/histograms/agri_data_hist.tex
\begin{tikzpicture}

\definecolor{darkgray176}{RGB}{176,176,176}
\definecolor{darkorange25512714}{RGB}{255,127,14}
\definecolor{forestgreen4416044}{RGB}{44,160,44}
\definecolor{lightgray204}{RGB}{204,204,204}
\definecolor{steelblue31119180}{RGB}{31,119,180}

\begin{axis}[
legend cell align={left},
legend style={fill opacity=0.8, draw opacity=1, text opacity=1, draw=lightgray204},
tick align=outside,
tick pos=left,
x grid style={darkgray176},
xmin=-0.975, xmax=14.975,
xtick style={color=black},
xtick={0.15,1.15,2.15,3.15,4.15,5.15,6.15,7.15,8.15,9.15,10.15,11.15,12.15,13.15,14.15},
xticklabel style={rotate=90.0},
xticklabels={
  whiptail,
  meerkat,
  koala,
  Gila monster,
  consomme,
  dung beetle,
  sidewinder,
  cliff,
  mongoose,
  pot,
  fiddler crab,
  coucal,
  water snake,
  polecat,
  wolf spider
},
y grid style={darkgray176},
ymin=0, ymax=8824.2,
ytick style={color=black}
]
\draw[draw=none,fill=steelblue31119180] (axis cs:-0.25,0) rectangle (axis cs:-0.05,8404);
\addlegendimage{ybar,ybar legend,draw=none,fill=steelblue31119180}
\addlegendentry{DNDD}

\draw[draw=none,fill=steelblue31119180] (axis cs:0.75,0) rectangle (axis cs:0.95,4898);
\draw[draw=none,fill=steelblue31119180] (axis cs:1.75,0) rectangle (axis cs:1.95,3959);
\draw[draw=none,fill=steelblue31119180] (axis cs:2.75,0) rectangle (axis cs:2.95,3498);
\draw[draw=none,fill=steelblue31119180] (axis cs:3.75,0) rectangle (axis cs:3.95,3253);
\draw[draw=none,fill=steelblue31119180] (axis cs:4.75,0) rectangle (axis cs:4.95,2947);
\draw[draw=none,fill=steelblue31119180] (axis cs:5.75,0) rectangle (axis cs:5.95,2323);
\draw[draw=none,fill=steelblue31119180] (axis cs:6.75,0) rectangle (axis cs:6.95,2176);
\draw[draw=none,fill=steelblue31119180] (axis cs:7.75,0) rectangle (axis cs:7.95,1868);
\draw[draw=none,fill=steelblue31119180] (axis cs:8.75,0) rectangle (axis cs:8.95,1816);
\draw[draw=none,fill=steelblue31119180] (axis cs:9.75,0) rectangle (axis cs:9.95,1784);
\draw[draw=none,fill=steelblue31119180] (axis cs:10.75,0) rectangle (axis cs:10.95,1713);
\draw[draw=none,fill=steelblue31119180] (axis cs:11.75,0) rectangle (axis cs:11.95,1292);
\draw[draw=none,fill=steelblue31119180] (axis cs:12.75,0) rectangle (axis cs:12.95,626);
\draw[draw=none,fill=steelblue31119180] (axis cs:13.75,0) rectangle (axis cs:13.95,593);
\draw[draw=none,fill=darkorange25512714] (axis cs:-0.1,0) rectangle (axis cs:0.1,4525);
\addlegendimage{ybar,ybar legend,draw=none,fill=darkorange25512714}
\addlegendentry{DDGAN}

\draw[draw=none,fill=darkorange25512714] (axis cs:0.9,0) rectangle (axis cs:1.1,4702);
\draw[draw=none,fill=darkorange25512714] (axis cs:1.9,0) rectangle (axis cs:2.1,1738);
\draw[draw=none,fill=darkorange25512714] (axis cs:2.9,0) rectangle (axis cs:3.1,3250);
\draw[draw=none,fill=darkorange25512714] (axis cs:3.9,0) rectangle (axis cs:4.1,5962);
\draw[draw=none,fill=darkorange25512714] (axis cs:4.9,0) rectangle (axis cs:5.1,1324);
\draw[draw=none,fill=darkorange25512714] (axis cs:5.9,0) rectangle (axis cs:6.1,3322);
\draw[draw=none,fill=darkorange25512714] (axis cs:6.9,0) rectangle (axis cs:7.1,3263);
\draw[draw=none,fill=darkorange25512714] (axis cs:7.9,0) rectangle (axis cs:8.1,1368);
\draw[draw=none,fill=darkorange25512714] (axis cs:8.9,0) rectangle (axis cs:9.1,2420);
\draw[draw=none,fill=darkorange25512714] (axis cs:9.9,0) rectangle (axis cs:10.1,983);
\draw[draw=none,fill=darkorange25512714] (axis cs:10.9,0) rectangle (axis cs:11.1,548);
\draw[draw=none,fill=darkorange25512714] (axis cs:11.9,0) rectangle (axis cs:12.1,711);
\draw[draw=none,fill=darkorange25512714] (axis cs:12.9,0) rectangle (axis cs:13.1,485);
\draw[draw=none,fill=darkorange25512714] (axis cs:13.9,0) rectangle (axis cs:14.1,1071);
\draw[draw=none,fill=forestgreen4416044] (axis cs:0.05,0) rectangle (axis cs:0.25,6112);
\addlegendimage{ybar,ybar legend,draw=none,fill=forestgreen4416044}
\addlegendentry{Proj.FastGAN}

\draw[draw=none,fill=forestgreen4416044] (axis cs:1.05,0) rectangle (axis cs:1.25,4343);
\draw[draw=none,fill=forestgreen4416044] (axis cs:2.05,0) rectangle (axis cs:2.25,2999);
\draw[draw=none,fill=forestgreen4416044] (axis cs:3.05,0) rectangle (axis cs:3.25,3741);
\draw[draw=none,fill=forestgreen4416044] (axis cs:4.05,0) rectangle (axis cs:4.25,2199);
\draw[draw=none,fill=forestgreen4416044] (axis cs:5.05,0) rectangle (axis cs:5.25,3202);
\draw[draw=none,fill=forestgreen4416044] (axis cs:6.05,0) rectangle (axis cs:6.25,2365);
\draw[draw=none,fill=forestgreen4416044] (axis cs:7.05,0) rectangle (axis cs:7.25,1647);
\draw[draw=none,fill=forestgreen4416044] (axis cs:8.05,0) rectangle (axis cs:8.25,1280);
\draw[draw=none,fill=forestgreen4416044] (axis cs:9.05,0) rectangle (axis cs:9.25,2046);
\draw[draw=none,fill=forestgreen4416044] (axis cs:10.05,0) rectangle (axis cs:10.25,1586);
\draw[draw=none,fill=forestgreen4416044] (axis cs:11.05,0) rectangle (axis cs:11.25,1078);
\draw[draw=none,fill=forestgreen4416044] (axis cs:12.05,0) rectangle (axis cs:12.25,909);
\draw[draw=none,fill=forestgreen4416044] (axis cs:13.05,0) rectangle (axis cs:13.25,713);
\draw[draw=none,fill=forestgreen4416044] (axis cs:14.05,0) rectangle (axis cs:14.25,825);
\end{axis}

\end{tikzpicture}

%% file: figures/packet_transform/lvl_3_packs_new.tex
\begin{tikzpicture}[thick,fill opacity=.3,text opacity=1]
    \draw[thick, fill=tabGreen] (22.06, 3.45) rectangle (22.86, 2.64) node[pos=.5] {\large {\bf{aaa}}};
    \draw[thick, fill=tabGreen] (22.92, 3.45) rectangle (23.72, 2.64) node[pos=.5] {\large {\bf{aa}}{\bf{h}}};
    \draw[thick, fill=tabGreen] (23.78, 3.45) rectangle (24.58, 2.64) node[pos=.5] {\large {\bf{a}}{\bf{h}}{\bf{a}}};
    \draw[thick, fill=tabGreen] (24.64, 3.45) rectangle (25.42, 2.64) node[pos=.5] {\large {\bf{a}}{\bf{h}}{\bf{h}}};
    \draw[thick, fill=tabBlue] (25.55, 3.45) rectangle (26.35, 2.64) node[pos=.5] {\large {\bf{h}}{\bf{aa}}};
    \draw[thick, fill=tabBlue] (26.41, 3.45) rectangle (27.21, 2.64) node[pos=.5] {\large {\bf{h}}{\bf{a}}{\bf{h}}};
    \draw[thick, fill=tabBlue] (27.27, 3.45) rectangle (28.07, 2.64) node[pos=.5] {\large {\bf{hh}}{\bf{a}}};
    \draw[thick, fill=tabBlue] (28.13, 3.45) rectangle (28.93, 2.64) node[pos=.5] {\large {\bf{hhh}}};

    \draw[thick, fill=tabGreen] (22.06, 2.58) rectangle (22.86, 1.79) node[pos=.5] {\large {\bf{aa}}{\bf{v}}};
    \draw[thick, fill=tabGreen] (22.92, 2.58) rectangle (23.72, 1.79) node[pos=.5] {\large {\bf{aa}}{\bf{d}}};
    \draw[thick, fill=tabGreen] (23.78, 2.58) rectangle (24.58, 1.79) node[pos=.5] {\large {\bf{a}}{\bf{h}}{\bf{v}}};
    \draw[thick, fill=tabGreen] (24.64, 2.58) rectangle (25.42, 1.79) node[pos=.5] {\large {\bf{a}}{\bf{h}}{\bf{d}}};
    \draw[thick, fill=tabBlue] (25.55, 2.58) rectangle (26.35, 1.79) node[pos=.5] {\large {\bf{h}}{\bf{a}}\bf{v}};
    \draw[thick, fill=tabBlue] (26.41, 2.58) rectangle (27.21, 1.79) node[pos=.5] {\large {\bf{h}}{\bf{a}}{\bf{d}}};
    \draw[thick, fill=tabBlue] (27.27, 2.58) rectangle (28.07, 1.79) node[pos=.5] {\large {\bf{hh}}\bf{v}};
    \draw[thick, fill=tabBlue] (28.13, 2.58) rectangle (28.93, 1.79) node[pos=.5] {\large {\bf{hh}}{\bf{d}}};

    \draw[thick, fill=tabGreen] (22.06, 1.74) rectangle (22.86, 0.93) node[pos=.5] {\large {\bf{a}}{\bf{v}}{\bf{a}}};
    \draw[thick, fill=tabGreen] (22.92, 1.74) rectangle (23.72, 0.93) node[pos=.5] {\large {\bf{a}}{\bf{v}}{\bf{h}}};
    \draw[thick, fill=tabGreen] (23.78, 1.74) rectangle (24.58, 0.93) node[pos=.5] {\large {\bf{a}}{\bf{d}}{\bf{a}}};
    \draw[thick, fill=tabGreen] (24.64, 1.74) rectangle (25.42, 0.93) node[pos=.5] {\large {\bf{a}}{\bf{d}}{\bf{h}}};
    \draw[thick, fill=tabBlue] (25.55, 1.74) rectangle (26.35, 0.93) node[pos=.5] {\large {\bf{h}}{\bf{v}}{\bf{a}}};
    \draw[thick, fill=tabBlue] (26.41, 1.74) rectangle (27.21, 0.93) node[pos=.5] {\large {\bf{h}}{\bf{v}}{\bf{h}}};
    \draw[thick, fill=tabBlue] (27.27, 1.74) rectangle (28.07, 0.93) node[pos=.5] {\large {\bf{h}}{\bf{d}}{\bf{a}}};
    \draw[thick, fill=tabBlue] (28.13, 1.74) rectangle (28.93, 0.93) node[pos=.5] {\large {\bf{h}}{\bf{d}}{\bf{h}}};

    \draw[thick, fill=tabGreen] (22.06, 0.87) rectangle (22.86, 0.05) node[pos=.5] {\large {\bf{a}}{\bf{v}}{\bf{v}}};
    \draw[thick, fill=tabGreen] (22.92, 0.87) rectangle (23.72, 0.05) node[pos=.5] {\large {\bf{a}}{\bf{v}}{\bf{d}}};
    \draw[thick, fill=tabGreen] (23.78, 0.87) rectangle (24.58, 0.05) node[pos=.5] {\large {\bf{a}}{\bf{d}}{\bf{v}}};
    \draw[thick, fill=tabGreen] (24.64, 0.87) rectangle (25.42, 0.05) node[pos=.5] {\large {\bf{a}}{\bf{d}}{\bf{d}}};
    \draw[thick, fill=tabBlue] (25.55, 0.87) rectangle (26.35, 0.05) node[pos=.5] {\large {\bf{h}}{\bf{v}}{\bf{v}}};
    \draw[thick, fill=tabBlue] (26.41, 0.87) rectangle (27.21, 0.05) node[pos=.5] {\large {\bf{h}}{\bf{v}}{\bf{d}}};
    \draw[thick, fill=tabBlue] (27.27, 0.87) rectangle (28.07, 0.05) node[pos=.5] {\large {\bf{h}}{\bf{d}}{\bf{v}}};
    \draw[thick, fill=tabBlue] (28.13, 0.87) rectangle (28.93, 0.05) node[pos=.5] {\large {\bf{h}}{\bf{d}}{\bf{d}}};

    \draw[thick, fill=tabGray] (22.06, -0.04) rectangle (22.86, -0.85) node[pos=.5] {\large {\bf{v}}{\bf{a}}{\bf{a}}};
    \draw[thick, fill=tabGray] (22.92, -0.04) rectangle (23.72, -0.85) node[pos=.5] {\large {\bf{v}}{\bf{a}}{\bf{h}}};
    \draw[thick, fill=tabGray] (23.78, -0.04) rectangle (24.58, -0.85) node[pos=.5] {\large {\bf{v}}{\bf{h}}{\bf{a}}};
    \draw[thick, fill=tabGray] (24.64, -0.04) rectangle (25.42, -0.85) node[pos=.5] {\large {\bf{v}}{\bf{h}}{\bf{h}}};
    \draw[thick, fill=tabPurple] (25.55, -0.04) rectangle (26.35, -0.85) node[pos=.5] {\large {\bf{d}}{\bf{aa}}};
    \draw[thick, fill=tabPurple] (26.41, -0.04) rectangle (27.21, -0.85) node[pos=.5] {\large {\bf{d}}{\bf{a}}{\bf{h}}};
    \draw[thick, fill=tabPurple] (27.27, -0.04) rectangle (28.07, -0.85) node[pos=.5] {\large {\bf{d}}{\bf{h}}{\bf{a}}};
    \draw[thick, fill=tabPurple] (28.13, -0.04) rectangle (28.93, -0.85) node[pos=.5] {\large {\bf{d}}{\bf{h}}{\bf{h}}};

    \draw[thick, fill=tabGray] (22.06, -0.91) rectangle (22.86, -1.72) node[pos=.5] {\large {\bf{v}}{\bf{a}}{\bf{v}}};
    \draw[thick, fill=tabGray] (22.92, -0.91) rectangle (23.72, -1.72) node[pos=.5] {\large {\bf{v}}{\bf{a}}{\bf{d}}};
    \draw[thick, fill=tabGray] (23.78, -0.91) rectangle (24.58, -1.72) node[pos=.5] {\large {\bf{v}}{\bf{h}}{\bf{v}}};
    \draw[thick, fill=tabGray] (24.64, -0.91) rectangle (25.42, -1.72) node[pos=.5] {\large {\bf{v}}{\bf{h}}{\bf{d}}};
    \draw[thick, fill=tabPurple] (25.55, -0.91) rectangle (26.35, -1.72) node[pos=.5] {\large {\bf{d}}{\bf{a}}{\bf{v}}};
    \draw[thick, fill=tabPurple] (26.41, -0.91) rectangle (27.21, -1.72) node[pos=.5] {\large {\bf{d}}{\bf{a}}{\bf{d}}};
    \draw[thick, fill=tabPurple] (27.27, -0.91) rectangle (28.07, -1.72) node[pos=.5] {\large {\bf{d}}{\bf{h}}{\bf{v}}};
    \draw[thick, fill=tabPurple] (28.13, -0.91) rectangle (28.93, -1.72) node[pos=.5] {\large {\bf{d}}{\bf{h}}{\bf{d}}};

    \draw[thick, fill=tabGray] (22.06, -1.78) rectangle (22.86, -2.59) node[pos=.5] {\large {\bf{v}}{\bf{v}}{\bf{a}}};
    \draw[thick, fill=tabGray] (22.92, -1.78) rectangle (23.72, -2.59) node[pos=.5] {\large {\bf{v}}{\bf{v}}{\bf{h}}};
    \draw[thick, fill=tabGray] (23.78, -1.78) rectangle (24.58, -2.59) node[pos=.5] {\large {\bf{v}}{\bf{d}}{\bf{a}}};
    \draw[thick, fill=tabGray] (24.64, -1.78) rectangle (25.42, -2.59) node[pos=.5] {\large {\bf{v}}{\bf{d}}{\bf{h}}};
    \draw[thick, fill=tabPurple] (25.55, -1.78) rectangle (26.35, -2.59) node[pos=.5] {\large {\bf{d}}{\bf{v}}{\bf{a}}};
    \draw[thick, fill=tabPurple] (26.41, -1.78) rectangle (27.21, -2.59) node[pos=.5] {\large {\bf{d}}{\bf{v}}{\bf{h}}};
    \draw[thick, fill=tabPurple] (27.27, -1.78) rectangle (28.07, -2.59) node[pos=.5] {\large {\bf{d}}{\bf{d}}{\bf{a}}};
    \draw[thick, fill=tabPurple] (28.13, -1.78) rectangle (28.93, -2.59) node[pos=.5] {\large {\bf{d}}{\bf{d}}{\bf{h}}};

    \draw[thick, fill=tabGray] (22.06, -2.65) rectangle (22.86, -3.45) node[pos=.5] {\large {\bf{v}}{\bf{v}}{\bf{v}}};
    \draw[thick, fill=tabGray] (22.92, -2.65) rectangle (23.72, -3.45) node[pos=.5] {\large {\bf{v}}{\bf{v}}{\bf{d}}};
    \draw[thick, fill=tabGray] (23.78, -2.65) rectangle (24.58, -3.45) node[pos=.5] {\large {\bf{v}}{\bf{d}}{\bf{v}}};
    \draw[thick, fill=tabGray] (24.64, -2.65) rectangle (25.42, -3.45) node[pos=.5] {\large {\bf{v}}{\bf{d}}{\bf{d}}};
    \draw[thick, fill=tabPurple] (25.55, -2.65) rectangle (26.35, -3.45) node[pos=.5] {\large {\bf{d}}{\bf{v}}{\bf{v}}};
    \draw[thick, fill=tabPurple] (26.41, -2.65) rectangle (27.21, -3.45) node[pos=.5] {\large {\bf{d}}{\bf{v}}{\bf{d}}};
    \draw[thick, fill=tabPurple] (27.27, -2.65) rectangle (28.07, -3.45) node[pos=.5] {\large {\bf{d}}{\bf{d}}{\bf{v}}};
    \draw[thick, fill=tabPurple] (28.13, -2.65) rectangle (28.93, -3.45) node[pos=.5] {\large {\bf{d}}{\bf{d}}{\bf{d}}};
\end{tikzpicture}

%% file: figures/corrupt_plots/blur.tex
\begin{tikzpicture}

\definecolor{crimson2143940}{RGB}{214,39,40}
\definecolor{darkgray176}{RGB}{176,176,176}
\definecolor{darkorange25512714}{RGB}{255,127,14}
\definecolor{forestgreen4416044}{RGB}{44,160,44}
\definecolor{lightgray204}{RGB}{204,204,204}
\definecolor{steelblue31119180}{RGB}{31,119,180}

\begin{axis}[
legend cell align={left},
legend style={
  fill opacity=0.8,
  draw opacity=1,
  text opacity=1,
  at={(0.5,0.45)},
  anchor=north west,
  draw=lightgray204
},
tick align=outside,
tick pos=left,
x grid style={darkgray176},
xlabel={Blur intensity},
xmajorgrids,
xmin=-0.5, xmax=10.5,
xtick style={color=black},
y grid style={darkgray176},
ylabel={normalized cost},
ymajorgrids,
ymin=-0.05, ymax=1.05,
ytick style={color=black}
]
\addplot [semithick, steelblue31119180]
table {%
0 3.48292730669284e-08
0.204081632653061 0
0.408163265306122 0.0193172693626887
0.612244897959184 0.239686125172065
0.816326530612245 0.391480381812028
1.02040816326531 0.496009923713217
1.22448979591837 0.564542982757312
1.42857142857143 0.615169455044115
1.63265306122449 0.655321952891906
1.83673469387755 0.688623992605604
2.04081632653061 0.711160441832363
2.24489795918367 0.717290140460438
2.44897959183673 0.729567143366432
2.6530612244898 0.745000995625431
2.85714285714286 0.747523888183091
3.06122448979592 0.736744668398138
3.26530612244898 0.724244695807196
3.46938775510204 0.71219883001476
3.6734693877551 0.701416903415985
3.87755102040816 0.696609248352293
4.08163265306122 0.694620358032951
4.28571428571429 0.692113700735633
4.48979591836735 0.690185173548215
4.69387755102041 0.692204192331733
4.89795918367347 0.696765252646025
5.10204081632653 0.703129067620514
5.30612244897959 0.711678438025479
5.51020408163265 0.719164552145759
5.71428571428571 0.728075498452845
5.91836734693878 0.739240376462439
6.12244897959184 0.750310491695736
6.3265306122449 0.76217802293237
6.53061224489796 0.776887740603425
6.73469387755102 0.790548545216151
6.93877551020408 0.805099603856118
7.14285714285714 0.821363074904736
7.3469387755102 0.838458395325417
7.55102040816327 0.852691862810417
7.75510204081633 0.865571263931592
7.95918367346939 0.877707183101094
8.16326530612245 0.8893640922806
8.36734693877551 0.902365484175937
8.57142857142857 0.914486715912426
8.77551020408163 0.925383986552773
8.97959183673469 0.936842395601037
9.18367346938776 0.949220413983345
9.38775510204082 0.960644877540554
9.59183673469388 0.972837586631826
9.79591836734694 0.985486591238554
10 1
};
\addlegendentry{FID}
\addplot [semithick, darkorange25512714]
table {%
0 2.68674831905241e-07
0.204081632653061 0
0.408163265306122 0.00887367419147819
0.612244897959184 0.119695140822222
0.816326530612245 0.240990707993594
1.02040816326531 0.340623890670366
1.22448979591837 0.429453558805307
1.42857142857143 0.509500369887369
1.63265306122449 0.579707574715972
1.83673469387755 0.638853653971063
2.04081632653061 0.686645424394861
2.24489795918367 0.723831731725754
2.44897959183673 0.752008596450698
2.6530612244898 0.773133569883891
2.85714285714286 0.78909513341206
3.06122448979592 0.801515311759521
3.26530612244898 0.811589186475002
3.46938775510204 0.820173484627944
3.6734693877551 0.827843594053633
3.87755102040816 0.834970512122494
4.08163265306122 0.841775469210011
4.28571428571429 0.848416897991681
4.48979591836735 0.854961106264199
4.69387755102041 0.861448466793742
4.89795918367347 0.867896132785388
5.10204081632653 0.874298459586898
5.30612244897959 0.880670255993482
5.51020408163265 0.886992210646526
5.71428571428571 0.893253907211652
5.91836734693878 0.899444628595018
6.12244897959184 0.905545983731198
6.3265306122449 0.911565993246973
6.53061224489796 0.917487005085118
6.73469387755102 0.923302107792946
6.93877551020408 0.929005583445443
7.14285714285714 0.934592888519485
7.3469387755102 0.940054533718766
7.55102040816327 0.945400822306533
7.75510204081633 0.950623187257596
7.95918367346939 0.95572075954692
8.16326530612245 0.960693306425877
8.36734693877551 0.965535988728682
8.57142857142857 0.970260638068999
8.77551020408163 0.974862602207204
8.97959183673469 0.979343434842433
9.18367346938776 0.983705026564498
9.38775510204082 0.987949442043805
9.59183673469388 0.99207500273433
9.79591836734694 0.996092497050832
10 1
};
\addlegendentry{FWD}
\addplot [semithick, forestgreen4416044]
table {%
0 3.64464337795104e-08
0.204081632653061 0
0.408163265306122 0.132145930548594
0.612244897959184 0.531534033840297
0.816326530612245 0.811043688414261
1.02040816326531 0.93250931489699
1.22448979591837 0.963884057639857
1.42857142857143 0.975667327785435
1.63265306122449 0.982864644513107
1.83673469387755 0.986763092470705
2.04081632653061 0.989084547710697
2.24489795918367 0.990668994190264
2.44897959183673 0.991821875010314
2.6530612244898 0.992696495820947
2.85714285714286 0.993383391572229
3.06122448979592 0.993941961392531
3.26530612244898 0.994408157641292
3.46938775510204 0.994808216566125
3.6734693877551 0.995159774478172
3.87755102040816 0.995474549476924
4.08163265306122 0.995760754308926
4.28571428571429 0.996024765748346
4.48979591836735 0.996271239202223
4.69387755102041 0.996501103640186
4.89795918367347 0.9967178439817
5.10204081632653 0.99692076276978
5.30612244897959 0.997113941763889
5.51020408163265 0.997297001893114
5.71428571428571 0.997470970281955
5.91836734693878 0.997636369473777
6.12244897959184 0.997792916050332
6.3265306122449 0.997943914141989
6.53061224489796 0.998087433762524
6.73469387755102 0.998225565892263
6.93877551020408 0.998358345849145
7.14285714285714 0.998486656694256
7.3469387755102 0.9986092263928
7.55102040816327 0.998729051215925
7.75510204081633 0.99884592486431
7.95918367346939 0.998959588331108
8.16326530612245 0.99907085279236
8.36734693877551 0.999178864275538
8.57142857142857 0.999285769926613
8.77551020408163 0.999390998996219
8.97959183673469 0.999495258412485
9.18367346938776 0.99959813524755
9.38775510204082 0.999699756973562
9.59183673469388 0.999800180242443
9.79591836734694 0.999900405484884
10 1
};
\addlegendentry{FD-DINOv2}
\end{axis}

\end{tikzpicture}

%% file: figures/corrupt_plots/uniform.tex
\begin{tikzpicture}

\definecolor{crimson2143940}{RGB}{214,39,40}
\definecolor{darkgray176}{RGB}{176,176,176}
\definecolor{darkorange25512714}{RGB}{255,127,14}
\definecolor{forestgreen4416044}{RGB}{44,160,44}
\definecolor{lightgray204}{RGB}{204,204,204}
\definecolor{steelblue31119180}{RGB}{31,119,180}

\begin{axis}[
legend cell align={left},
legend style={
  fill opacity=0.8,
  draw opacity=1,
  text opacity=1,
  at={(0.03,0.97)},
  anchor=north west,
  draw=lightgray204
},
tick align=outside,
tick pos=left,
x grid style={darkgray176},
xlabel={Uniform noise intensity},
xmajorgrids,
xmin=-0.05, xmax=1.05,
xtick style={color=black},
y grid style={darkgray176},
ylabel={normalized cost},
ymajorgrids,
ymin=-0.05, ymax=1.05,
ytick style={color=black}
]
\addplot [semithick, steelblue31119180]
table {%
0 0
0.0204081632653061 0.0129757922192263
0.0408163265306122 0.0475444795545931
0.0612244897959184 0.0897133454460835
0.0816326530612245 0.116516519641117
0.102040816326531 0.132421344310259
0.122448979591837 0.15439046913939
0.142857142857143 0.166942243128832
0.163265306122449 0.185185586669923
0.183673469387755 0.207113661347152
0.204081632653061 0.215987017104103
0.224489795918367 0.223740854646656
0.244897959183673 0.251570101681561
0.26530612244898 0.261171384267914
0.285714285714286 0.285119564666334
0.306122448979592 0.285406620410633
0.326530612244898 0.302199597474875
0.346938775510204 0.321531106498088
0.36734693877551 0.339510983137457
0.387755102040816 0.377582431807132
0.408163265306122 0.397568588382709
0.428571428571429 0.443515096826458
0.448979591836735 0.458172438545675
0.469387755102041 0.499235967658158
0.489795918367347 0.544350922934311
0.510204081632653 0.573936659334837
0.530612244897959 0.676484969750628
0.551020408163265 0.676752666112198
0.571428571428571 0.765608569019494
0.591836734693878 0.830589878962999
0.612244897959184 0.828468177272103
0.63265306122449 0.930498574634488
0.653061224489796 0.890739439415566
0.673469387755102 0.980880646029978
0.693877551020408 0.921159766136068
0.714285714285714 0.945207329938097
0.73469387755102 0.884803895092591
0.755102040816326 0.881434811074286
0.775510204081633 0.893685337778236
0.795918367346939 0.88150820443952
0.816326530612245 0.88605050498568
0.836734693877551 0.89693187524232
0.857142857142857 0.896709448844054
0.877551020408163 0.925254596083571
0.897959183673469 0.930362542210628
0.918367346938775 0.953679851347832
0.938775510204082 0.97558548066564
0.959183673469388 0.965399337380886
0.979591836734694 1
1 0.982552225796633
};
\addlegendentry{FID}
\addplot [semithick, darkorange25512714]
table {%
0 0
0.0204081632653061 0.000424629824753144
0.0408163265306122 0.00169344125837941
0.0612244897959184 0.00383702955614945
0.0816326530612245 0.00681472405890793
0.102040816326531 0.0106335853238471
0.122448979591837 0.0152939742947015
0.142857142857143 0.0207864663812664
0.163265306122449 0.0272450237554451
0.183673469387755 0.0343885101660131
0.204081632653061 0.0426486983085596
0.224489795918367 0.0514123874070744
0.244897959183673 0.061470431198961
0.26530612244898 0.0717986467116432
0.285714285714286 0.0829397297003761
0.306122448979592 0.0958806704048536
0.326530612244898 0.108749414347232
0.346938775510204 0.122672188628308
0.36734693877551 0.137349612084585
0.387755102040816 0.153412849285176
0.408163265306122 0.170226617538952
0.428571428571429 0.187285357413136
0.448979591836735 0.204795582705235
0.469387755102041 0.224235573518412
0.489795918367347 0.244140812010887
0.510204081632653 0.264960719705624
0.530612244897959 0.286647668941695
0.551020408163265 0.308051865041454
0.571428571428571 0.332442884406125
0.591836734693878 0.355127414992481
0.612244897959184 0.379824251684121
0.63265306122449 0.406329486753645
0.653061224489796 0.433350857837929
0.673469387755102 0.46063587323768
0.693877551020408 0.489463691771177
0.714285714285714 0.518028427743395
0.73469387755102 0.547191264499908
0.755102040816326 0.580063622314797
0.775510204081633 0.612934626041849
0.795918367346939 0.640493580529347
0.816326530612245 0.675791346754805
0.836734693877551 0.709042626713612
0.857142857142857 0.745764377292228
0.877551020408163 0.777481836765541
0.897959183673469 0.819269878074264
0.918367346938775 0.854794582701938
0.938775510204082 0.893120077168181
0.959183673469388 0.930704272092514
0.979591836734694 0.969183960112465
1 1
};
\addlegendentry{FWD}
\addplot [semithick, forestgreen4416044]
table {%
0 0
0.0204081632653061 0.0106048726388023
0.0408163265306122 0.0376796660047194
0.0612244897959184 0.0771959101774722
0.0816326530612245 0.129891697166048
0.102040816326531 0.17933610926167
0.122448979591837 0.224571511605303
0.142857142857143 0.264052781953951
0.163265306122449 0.303590923265886
0.183673469387755 0.334192974601409
0.204081632653061 0.369289200319215
0.224489795918367 0.400002627413553
0.244897959183673 0.432586044757174
0.26530612244898 0.46404349407835
0.285714285714286 0.500839046119315
0.306122448979592 0.538698905304837
0.326530612244898 0.568553017614736
0.346938775510204 0.605384682398974
0.36734693877551 0.64245490302466
0.387755102040816 0.678011135787198
0.408163265306122 0.717537385466897
0.428571428571429 0.751206251526031
0.448979591836735 0.801264786625029
0.469387755102041 0.838300321535477
0.489795918367347 0.880008588823132
0.510204081632653 0.897002041647401
0.530612244897959 0.927610326592795
0.551020408163265 0.938993589351203
0.571428571428571 0.951201172156694
0.591836734693878 0.952987868243594
0.612244897959184 0.970155672309318
0.63265306122449 0.980439499911374
0.653061224489796 0.984205018639285
0.673469387755102 0.984787094354359
0.693877551020408 0.987412666196164
0.714285714285714 0.993270005140208
0.73469387755102 1
0.755102040816326 0.997610438067879
0.775510204081633 0.996426439190264
0.795918367346939 0.99737854295282
0.816326530612245 0.991138174827131
0.836734693877551 0.994271909205206
0.857142857142857 0.992449640892479
0.877551020408163 0.978534530735902
0.897959183673469 0.99127370121081
0.918367346938775 0.986824289739714
0.938775510204082 0.983020545167184
0.959183673469388 0.982295794499205
0.979591836734694 0.977866892170164
1 0.972272690478564
};
\addlegendentry{FD-DINOv2}
\end{axis}

\end{tikzpicture}

%% file: figures/corrupt_plots/pil.tex
\begin{tikzpicture}

\definecolor{crimson2143940}{RGB}{214,39,40}
\definecolor{darkgray176}{RGB}{176,176,176}
\definecolor{darkorange25512714}{RGB}{255,127,14}
\definecolor{forestgreen4416044}{RGB}{44,160,44}
\definecolor{lightgray204}{RGB}{204,204,204}
\definecolor{steelblue31119180}{RGB}{31,119,180}

\begin{axis}[
legend cell align={left},
legend style={fill opacity=0.8, draw opacity=1, text opacity=1, draw=lightgray204},
tick align=outside,
tick pos=left,
x grid style={darkgray176},
xlabel={PIL JPEG Quality},
xmajorgrids,
xmin=-3.9, xmax=103.9,
xtick style={color=black},
y grid style={darkgray176},
ylabel={normalized cost},
ymajorgrids,
ymin=-0.05, ymax=1.05,
ytick style={color=black}
]
\addplot [semithick, steelblue31119180]
table {%
1 1
2 0.998259362713728
3 0.953773036961347
4 0.602269165652017
5 0.431640653933713
6 0.362988479104234
7 0.331807279461929
8 0.2721878516951
9 0.244620352920296
10 0.218917197252971
11 0.201599446335112
12 0.193424331527836
13 0.168449956617696
14 0.166113331238172
15 0.164369454110422
16 0.149689443568234
17 0.132119564203068
18 0.127166329630186
19 0.117638072512516
20 0.118312070998691
21 0.111270217264591
22 0.109711376185661
23 0.103714206224177
24 0.0962535773323411
25 0.0888223642464135
26 0.0792689900244463
27 0.0765672360556402
28 0.0730444889243185
29 0.0691405476995955
30 0.0654737587208955
31 0.0633206059403861
32 0.0651611035405483
33 0.0613452478101814
34 0.0598551298152692
35 0.0620792452643046
36 0.0605384942440739
37 0.0589443766026651
38 0.0578863706035156
39 0.059260012782547
40 0.0583322714649663
41 0.0591627947784476
42 0.0621949278389521
43 0.0624326432061552
44 0.0623572778022809
45 0.0624424864812069
46 0.0629837279681991
47 0.0619449619107492
48 0.0579262040288946
49 0.0634456756755168
50 0.0609660017360281
51 0.0586588544168114
52 0.0585403011619815
53 0.0542069323576133
54 0.0530471592033162
55 0.0494913855715156
56 0.0496807341043066
57 0.0462896626405434
58 0.0410404252287671
59 0.0380850317809995
60 0.037364877948218
61 0.0308366111379761
62 0.029306773361616
63 0.0264338650143075
64 0.0223598871392978
65 0.0221034821298535
66 0.019171977212354
67 0.0183321245993372
68 0.0149343033801984
69 0.0135849347272667
70 0.0101699710344267
71 0.00684168410294476
72 0.00649357698563948
73 0.00334341108180121
74 0.000287447858703274
75 0
76 0.00199636486405292
77 0.00349838612740054
78 0.00432604649403063
79 0.00555053718982109
80 0.00728488905091426
81 0.00872409100709602
82 0.00991901048270275
83 0.0089276970814702
84 0.00806641307132181
85 0.00639387548676805
86 0.00441252925319986
87 0.00239009098215847
88 0.00129354586464164
89 0.00220673647302559
90 0.00306448624841499
91 0.00209518364679099
92 0.00182287592256145
93 0.00158573912189832
94 0.00175420639920886
95 0.00127902975660527
96 0.00117475914358693
97 0.00116971047746414
98 0.00107369163075271
99 0.000944788135085222
};
\addlegendentry{FID}
\addplot [semithick, darkorange25512714]
table {%
1 1
2 0.999704820621099
3 0.936316161162963
4 0.702836418443744
5 0.536356063074839
6 0.429759642925541
7 0.359044329332772
8 0.311605726807162
9 0.276122941142008
10 0.249298735264365
11 0.227345722000631
12 0.213099833920623
13 0.197123207801969
14 0.183557375419519
15 0.174845694810558
16 0.168191181748868
17 0.161016175997911
18 0.149945828267539
19 0.143430173513826
20 0.138101292943957
21 0.132923449168422
22 0.12953681212466
23 0.127425812302127
24 0.124070979668171
25 0.119728224632268
26 0.113094079113867
27 0.106676557133111
28 0.101987925400887
29 0.0989762017882265
30 0.0952435992901897
31 0.0914798697045056
32 0.0891092265956925
33 0.086397460961292
34 0.0855801475059585
35 0.0844824631362957
36 0.0830826767561736
37 0.0822207254026837
38 0.0808726813371653
39 0.0797101247873192
40 0.0789293795014526
41 0.0778609056415319
42 0.0777198667967133
43 0.0773594956440513
44 0.0770635732050683
45 0.076530516320349
46 0.0765338108877552
47 0.0764642628292936
48 0.075929623822772
49 0.0740890434820313
50 0.0713623164250342
51 0.0685472121999468
52 0.0621266918541365
53 0.0578394206323749
54 0.0538703382798731
55 0.0488905123124944
56 0.0443263471102647
57 0.0402790630775589
58 0.0365439440640205
59 0.0337115091287478
60 0.0303561294055813
61 0.0267431275008785
62 0.0243276732391651
63 0.0214994129595725
64 0.0193383299545977
65 0.0174666519857168
66 0.0153635927394211
67 0.0135376366812816
68 0.0114628179794573
69 0.00956405056847014
70 0.00748902483454132
71 0.00548083609612947
72 0.00436202170681205
73 0.00260359852215106
74 0.000372938837677006
75 0
76 0.00174499946571018
77 0.00308243526381548
78 0.003843651142757
79 0.00470552301596837
80 0.0057815134875319
81 0.00672233672441207
82 0.00757038699056751
83 0.00824920189991227
84 0.00702394391306269
85 0.00469458929407098
86 0.00280461323680994
87 0.00163810124103108
88 0.00149769597560523
89 0.00237178580333795
90 0.00309914888409979
91 0.00163060303568434
92 0.00134932102521007
93 0.00164730606560313
94 0.00132011155315042
95 0.00110331449820082
96 0.000990500998632399
97 0.000919967543308646
98 0.000852062993851017
99 0.000763386259185965
};
\addlegendentry{FWD}
\addplot [semithick, forestgreen4416044]
table {%
1 1
2 0.996033065175042
3 0.68981370794283
4 0.418412342877387
5 0.3226366220376
6 0.243395093005561
7 0.215625048013216
8 0.131848127853605
9 0.0968217810664445
10 0.056593493939025
11 0.0555309890929697
12 0.0719707221573088
13 0.0340760475855448
14 0.0359334330773368
15 0.0420784947635808
16 0.0410276063442316
17 0.0204970670698377
18 0.0124540972508856
19 0.0113289759542371
20 0.0195389468225279
21 0.0273156629740353
22 0.0221849483376478
23 0.0233534424577721
24 0.0236163526560033
25 0.0103921736921387
26 0.020412258507596
27 0.014079698917892
28 0.0126307038822732
29 0.0105289117162523
30 0.00608230898847092
31 0.00742688310197898
32 0.0051585329159923
33 0.00772089992108631
34 0.0101891242804351
35 0.0121402538833043
36 0.010086885528618
37 0.00863295300075945
38 0.00770783581410744
39 0.00869424397231214
40 0.00918619424120103
41 0.0119545267366427
42 0.0115696023417481
43 0.0115130627623118
44 0.0148591882171648
45 0.014553775735047
46 0.0158690996733933
47 0.0139175946051522
48 0.0110963888844166
49 0.0072429515715295
50 0.00901578617418871
51 0.00920835503546113
52 0.0166947508934536
53 0.0105276169483673
54 0.00982669223567562
55 0.00761708894939098
56 0.00633715827580585
57 0.00551398135668614
58 0.00448360239594954
59 0.00468126972621997
60 0.00437911547940669
61 0.00412887756914251
62 0.00401366162999943
63 0.0025650246041125
64 0.00288049318482449
65 0.00220987131261838
66 0.00183971704217383
67 0.00189746539331453
68 0.00283884491684569
69 0.00240074648877864
70 0.00197688263244754
71 0.00250251824433647
72 0.0016794535266887
73 0.00101111423944356
74 0
75 1.02805664010705e-05
76 0.000326326728262136
77 0.00214679100106894
78 0.00218033887202508
79 0.00231040250700759
80 0.00181569760224547
81 0.00180362557858135
82 0.00152098773508576
83 0.000984691570025981
84 0.00085443993607902
85 0.000621196006293827
86 0.00060162959216248
87 0.00043241922954337
88 0.000402456035230372
89 0.000689738415024901
90 0.000681518048566134
91 0.000458114253784143
92 0.000487487875713977
93 0.000400583740440353
94 0.000514539336075082
95 0.000427604803937874
96 0.000466230257825975
97 0.000444575616411911
98 0.000385983459995908
99 0.000335580579657243

};
\addlegendentry{FD-DINOv2}
\end{axis}

\end{tikzpicture}

%% file: figures/supplementary/fwt.tex
\begin{tikzpicture}
  \node[draw, rectangle] at (0,0) (x){$\mathbf{x}$};

  \node[draw, rectangle] at (1,  1)   (hl){$\mathbf{H_\mathcal{L}}\; \downarrow_2$};
  \node[draw, rectangle] at (2,  1.75) (hhh){$\mathbf{H_\mathcal{L}} \; \downarrow_2$};
  \node[draw, rectangle] at (2,  0.25) (hll){$\mathbf{H_\mathcal{H}} \; \downarrow_2$};

  \node[draw, rectangle] at (1,  -1.) (hh){$\mathbf{H_\mathcal{H}}\; \downarrow_2$ };
  \node[draw, rectangle] at (3.4, -1.0) (b0){$\mathbf{c}_{\mathcal{H},1}$};
  \node[draw, rectangle] at (3.4,  0.25) (b2){$\mathbf{c}_{\mathcal{H},2}$};
  \node[draw, rectangle] at (3.4,  1.75) (b3){$\mathbf{c}_{\mathcal{L},2}$};

  \node[draw, rectangle] at (5,  1.75)  (fhh){$\uparrow_2 \; \mathbf{F_\mathcal{L}}$};
  \node[draw, rectangle] at (5,  0.25)  (fll){$\uparrow_2 \; \mathbf{F_\mathcal{H}}$};

  \node[draw, rectangle] at (6, -1.0) (fh){$\uparrow_2 \; \mathbf{F_\mathcal{H}}$};
  \node[draw, rectangle] at (6,  1.) (fl){$\uparrow_2 \; \mathbf{F_\mathcal{L}}$};
  \node[draw, rectangle] at (7, 0) (xhat){$\hat{\mathbf{x}}$};

  \draw[->] (x) |- (hh);
  \draw[->] (x) |- (hl);
  \draw[->] (hl) |- (hhh);
  \draw[->] (hl) |- (hll);

  \draw[->] (hhh) -- (b3);
  \draw[->] (hll) -- (b2);
  \draw[->] (b2) -- (fll);
  \draw[->] (hh) -- (b0);
  \draw[->] (b0) -- (fh);

  \draw[dotted, ->] (b3) -- (fhh);

  \draw[->] (fll) -| (fl);
  \draw[->] (fhh) -| (fl);

  \draw[->] (fl) -| (xhat);
  \draw[->] (fh) -| (xhat);

\end{tikzpicture}

%% file: figures/supplementary/packets_1d.tex
\begin{tikzpicture}
  \node[draw, rectangle] at (0,0) (x){$\mathbf{x}$};

  \node[draw, rectangle] at (1,   2)   (hl){$\mathbf{H_\mathcal{L}}\; \downarrow_2$};
  \node[draw, rectangle] at (2,  3) (hhh){$\mathbf{H_\mathcal{L}} \; \downarrow_2$};
  \node[draw, rectangle] at (2,  1) (hll){$\mathbf{H_\mathcal{H}} \; \downarrow_2$};

  \node[draw, rectangle] at (1,  -2.) (hh){$\mathbf{H_\mathcal{H}}\; \downarrow_2$ };
  \node[draw, rectangle] at (2, -1) (hph){$\mathbf{H_\mathcal{L}} \downarrow_2$};
  \node[draw, rectangle] at (2, -3) (hpl){$\mathbf{H_\mathcal{H}} \downarrow 2$};

  \node[draw, rectangle] at (4, -3) (b0){$\mathbf{c}_{\mathcal{H},\mathcal{H}}$};
  \node[draw, rectangle] at (4, -1) (b1){$\mathbf{c}_{\mathcal{H},\mathcal{L}}$};
  \node[draw, rectangle] at (4,  1) (b2){$\mathbf{c}_{\mathcal{L},\mathcal{H}}$};
  \node[draw, rectangle] at (4,  3) (b3){$\mathbf{c}_{\mathcal{L},\mathcal{L}}$};

  \node[draw, rectangle] at (6, 3)  (fhh){$\uparrow_2 \; \mathbf{F_\mathcal{L}}$};
  \node[draw, rectangle] at (6, 1)  (fll){$\uparrow_2 \; \mathbf{F_\mathcal{H}} $};
  \node[draw, rectangle] at (6, -1) (fph){$\uparrow_2 \; \mathbf{F_\mathcal{L}} $};
  \node[draw, rectangle] at (6, -3) (fpl){$\uparrow_2 \; \mathbf{F_\mathcal{H}} $};

  \node[draw, rectangle] at (7, -2.0) (fh){$\uparrow_2 \; \mathbf{F_\mathcal{H}}$ };
  \node[draw, rectangle] at (7,  2.) (fl){$\uparrow_2 \; \mathbf{F_\mathcal{L}}$};
  \node[draw, rectangle] at (8, 0) (xhat){$\hat{\mathbf{x}}$};

  \draw[->] (x) |- (hh);
  \draw[->] (x) |- (hl);
  \draw[->] (hl) |- (hhh);
  \draw[->] (hl) |- (hll);
  \draw[->] (hh) |- (hph);
  \draw[->] (hh) |- (hpl);

  \draw[->] (hhh) -- (b3);
  \draw[->] (hll) -- (b2);
  \draw[->] (hph) -- (b1);
  \draw[->] (hpl) -- (b0);

  \draw[dotted, ->] (b3) -- (fhh);
  \draw[dotted, ->] (b2) -- (fll);
  \draw[dotted, ->] (b1) -- (fph);
  \draw[dotted, ->] (b0) -- (fpl);

  \draw[->] (fll) -| (fl);
  \draw[->] (fhh) -| (fl);
  \draw[->] (fpl) -| (fh);
  \draw[->] (fph) -| (fh);

  \draw[->] (fl) -| (xhat);
  \draw[->] (fh) -| (xhat);

\end{tikzpicture}

%% file: figures/supplementary/packets_2d.tex
\begin{tikzpicture}
  \node[draw, rectangle] at (0,0)   (I){$\mathbf{X}$};

  \node[draw, rectangle] at (-4.25, -1) (Ha){$\mathbf{H_a}\downarrow_2$};
  \node[draw, rectangle] at (-1, -1) (Hz){$\mathbf{H_h}\downarrow_2$};
  \node[draw, rectangle] at ( 1, -1) (Hv){$\mathbf{H_v}\downarrow_2$};
  \node[draw, rectangle] at ( 2.5, -1) (Hd){$\mathbf{H_d}\downarrow_2$};

  \draw[->] (I) -| (Ha);
  \draw[->] (I) -| (Hz);
  \draw[->] (I) -| (Hv);
  \draw[->] (I) -| (Hd);

  \node[draw, rectangle] at (-4.25, -2) (a){$\mathbf{p}_a$};
  \node[draw, rectangle] at (-1, -2) (z){$\mathbf{p}_h$};
  \node[draw, rectangle] at ( 1, -2) (v){$\mathbf{p}_v$};
  \node[draw, rectangle] at ( 2.5, -2) (d){$\mathbf{p}_d$};

  \draw[->] (Ha) -- (a);
  \draw[->] (Hz) -- (z);
  \draw[->] (Hv) -- (v);
  \draw[->] (Hd) -- (d);

  \node[draw, rectangle] at (-6.5, -3) (Haa){$\mathbf{H_a} \downarrow_2$};
  \node[draw, rectangle] at (-5, -3) (Haz){$\mathbf{H_h} \downarrow_2$};
  \node[draw, rectangle] at (-3.5, -3) (Hav){$\mathbf{H_v} \downarrow_2$};
  \node[draw, rectangle] at (-2, -3) (Had){$\mathbf{H_d} \downarrow_2$};

  \node[draw, rectangle] at (-6.5,  -3.75) (aa){$\mathbf{p}_{a,a}$};
  \node[draw, rectangle] at (-5,  -3.75) (az){$\mathbf{p}_{a,h}$};
  \node[draw, rectangle] at (-3.5,-3.75) (av){$\mathbf{p}_{a,v}$};
  \node[draw, rectangle] at (-2,  -3.75) (ad){$\mathbf{p}_{a,d}$};

  \draw[->] (Haa) -- (aa);
  \draw[->] (Haz) -- (az);
  \draw[->] (Hav) -- (av);
  \draw[->] (Had) -- (ad);

  \draw[->] (a) -| (Haa);
  \draw[->] (a) -| (Haz);
  \draw[->] (a) -| (Hav);
  \draw[->] (a) -| (Had);

  \node[draw, rectangle] at (-6.5,  -6) (Faa){$ \uparrow_2 \mathbf{F_a}$};
  \node[draw, rectangle] at (-5, -6) (Faz){$ \uparrow_2 \mathbf{F_h}$  };
  \node[draw, rectangle] at (-3.5, -6) (Fav){$ \uparrow_2 \mathbf{F_v}$};
  \node[draw, rectangle] at (-2,   -6) (Fad){$ \uparrow_2 \mathbf{F_d}$};
  
  \draw[->, dashed] (aa) -- (Faa);
  \draw[->, dashed] (az) -- (Faz);
  \draw[->, dashed] (av) -- (Fav);
  \draw[->, dashed] (ad) -- (Fad);

  \node[draw, rectangle] at (-4.25,-7) (Fa){$\uparrow_2 \mathbf{F_a}  $};
  \node[draw, rectangle] at (-1,  -7) (Fz){$\uparrow_2 \mathbf{F_h} $};
  \node[draw, rectangle] at ( 1,  -7) (Fv){$\uparrow_2 \mathbf{F_v} $};
  \node[draw, rectangle] at ( 2.5,  -7) (Fd){$\uparrow_2 \mathbf{F_d} $};

  \draw[->] (Faa) |- (Fa);
  \draw[->] (Faz) |- (Fa);
  \draw[->] (Fav) |- (Fa);
  \draw[->] (Fad) |- (Fa);

  \draw[->, dashed] (z) -- (Fz);
  \draw[->, dashed] (v) -- (Fv);
  \draw[->, dashed] (d) -- (Fd);

  \node[draw, rectangle] at (0,-8)   (hI){$\hat{\mathbf{X}}$};

  \draw[->] (Fa) |- (hI);
  \draw[->] (Fz) |- (hI);
  \draw[->] (Fv) |- (hI);
  \draw[->] (Fd) |- (hI);

\end{tikzpicture}

%% file: figures/histograms/celebahq_hist.tex
\begin{tikzpicture}

\definecolor{darkgray176}{RGB}{176,176,176}
\definecolor{darkorange25512714}{RGB}{255,127,14}
\definecolor{forestgreen4416044}{RGB}{44,160,44}
\definecolor{lightgray204}{RGB}{204,204,204}
\definecolor{steelblue31119180}{RGB}{31,119,180}

\begin{axis}[
legend cell align={left},
legend style={fill opacity=0.8, draw opacity=1, text opacity=1, draw=lightgray204},
tick align=outside,
tick pos=left,
x grid style={darkgray176},
xmin=-0.975, xmax=14.975,
xtick style={color=black},
xtick={0.15,1.15,2.15,3.15,4.15,5.15,6.15,7.15,8.15,9.15,10.15,11.15,12.15,13.15,14.15},
xticklabel style={rotate=90.0},
xticklabels={
  wig,
  suit,
  brassiere,
  bikini,
  neck brace,
  tench,
  lab coat,
  bow tie,
  fur coat,
  oboe,
  seat belt,
  lipstick,
  ping-pong ball,
  stethoscope,
  hair spray
},
y grid style={darkgray176},
ymin=0, ymax=10149.3,
ytick style={color=black}
]
\draw[draw=none,fill=steelblue31119180] (axis cs:-0.25,0) rectangle (axis cs:-0.05,6103);
\addlegendimage{ybar,ybar legend,draw=none,fill=steelblue31119180}
\addlegendentry{CelebAHQ}

\draw[draw=none,fill=steelblue31119180] (axis cs:0.75,0) rectangle (axis cs:0.95,3753);
\draw[draw=none,fill=steelblue31119180] (axis cs:1.75,0) rectangle (axis cs:1.95,3303);
\draw[draw=none,fill=steelblue31119180] (axis cs:2.75,0) rectangle (axis cs:2.95,2979);
\draw[draw=none,fill=steelblue31119180] (axis cs:3.75,0) rectangle (axis cs:3.95,943);
\draw[draw=none,fill=steelblue31119180] (axis cs:4.75,0) rectangle (axis cs:4.95,713);
\draw[draw=none,fill=steelblue31119180] (axis cs:5.75,0) rectangle (axis cs:5.95,687);
\draw[draw=none,fill=steelblue31119180] (axis cs:6.75,0) rectangle (axis cs:6.95,682);
\draw[draw=none,fill=steelblue31119180] (axis cs:7.75,0) rectangle (axis cs:7.95,633);
\draw[draw=none,fill=steelblue31119180] (axis cs:8.75,0) rectangle (axis cs:8.95,583);
\draw[draw=none,fill=steelblue31119180] (axis cs:9.75,0) rectangle (axis cs:9.95,451);
\draw[draw=none,fill=steelblue31119180] (axis cs:10.75,0) rectangle (axis cs:10.95,421);
\draw[draw=none,fill=steelblue31119180] (axis cs:11.75,0) rectangle (axis cs:11.95,336);
\draw[draw=none,fill=steelblue31119180] (axis cs:12.75,0) rectangle (axis cs:12.95,285);
\draw[draw=none,fill=steelblue31119180] (axis cs:13.75,0) rectangle (axis cs:13.95,276);
\draw[draw=none,fill=darkorange25512714] (axis cs:-0.1,0) rectangle (axis cs:0.1,6145);
\addlegendimage{ybar,ybar legend,draw=none,fill=darkorange25512714}
\addlegendentry{DDPM}

\draw[draw=none,fill=darkorange25512714] (axis cs:0.9,0) rectangle (axis cs:1.1,1759);
\draw[draw=none,fill=darkorange25512714] (axis cs:1.9,0) rectangle (axis cs:2.1,5945);
\draw[draw=none,fill=darkorange25512714] (axis cs:2.9,0) rectangle (axis cs:3.1,4511);
\draw[draw=none,fill=darkorange25512714] (axis cs:3.9,0) rectangle (axis cs:4.1,1257);
\draw[draw=none,fill=darkorange25512714] (axis cs:4.9,0) rectangle (axis cs:5.1,971);
\draw[draw=none,fill=darkorange25512714] (axis cs:5.9,0) rectangle (axis cs:6.1,592);
\draw[draw=none,fill=darkorange25512714] (axis cs:6.9,0) rectangle (axis cs:7.1,319);
\draw[draw=none,fill=darkorange25512714] (axis cs:7.9,0) rectangle (axis cs:8.1,1015);
\draw[draw=none,fill=darkorange25512714] (axis cs:8.9,0) rectangle (axis cs:9.1,815);
\draw[draw=none,fill=darkorange25512714] (axis cs:9.9,0) rectangle (axis cs:10.1,186);
\draw[draw=none,fill=darkorange25512714] (axis cs:10.9,0) rectangle (axis cs:11.1,92);
\draw[draw=none,fill=darkorange25512714] (axis cs:11.9,0) rectangle (axis cs:12.1,273);
\draw[draw=none,fill=darkorange25512714] (axis cs:12.9,0) rectangle (axis cs:13.1,90);
\draw[draw=none,fill=darkorange25512714] (axis cs:13.9,0) rectangle (axis cs:14.1,225);
\draw[draw=none,fill=forestgreen4416044] (axis cs:0.05,0) rectangle (axis cs:0.25,9666);
\addlegendimage{ybar,ybar legend,draw=none,fill=forestgreen4416044}
\addlegendentry{StyleSwin}

\draw[draw=none,fill=forestgreen4416044] (axis cs:1.05,0) rectangle (axis cs:1.25,5392);
\draw[draw=none,fill=forestgreen4416044] (axis cs:2.05,0) rectangle (axis cs:2.25,4199);
\draw[draw=none,fill=forestgreen4416044] (axis cs:3.05,0) rectangle (axis cs:3.25,4861);
\draw[draw=none,fill=forestgreen4416044] (axis cs:4.05,0) rectangle (axis cs:4.25,917);
\draw[draw=none,fill=forestgreen4416044] (axis cs:5.05,0) rectangle (axis cs:5.25,401);
\draw[draw=none,fill=forestgreen4416044] (axis cs:6.05,0) rectangle (axis cs:6.25,1085);
\draw[draw=none,fill=forestgreen4416044] (axis cs:7.05,0) rectangle (axis cs:7.25,2893);
\draw[draw=none,fill=forestgreen4416044] (axis cs:8.05,0) rectangle (axis cs:8.25,1339);
\draw[draw=none,fill=forestgreen4416044] (axis cs:9.05,0) rectangle (axis cs:9.25,429);
\draw[draw=none,fill=forestgreen4416044] (axis cs:10.05,0) rectangle (axis cs:10.25,2098);
\draw[draw=none,fill=forestgreen4416044] (axis cs:11.05,0) rectangle (axis cs:11.25,2321);
\draw[draw=none,fill=forestgreen4416044] (axis cs:12.05,0) rectangle (axis cs:12.25,528);
\draw[draw=none,fill=forestgreen4416044] (axis cs:13.05,0) rectangle (axis cs:13.25,657);
\draw[draw=none,fill=forestgreen4416044] (axis cs:14.05,0) rectangle (axis cs:14.25,734);
\end{axis}

\end{tikzpicture}

%% file: figures/histograms/lsun_church_hist.tex
\begin{tikzpicture}

\definecolor{darkgray176}{RGB}{176,176,176}
\definecolor{darkorange25512714}{RGB}{255,127,14}
\definecolor{forestgreen4416044}{RGB}{44,160,44}
\definecolor{lightgray204}{RGB}{204,204,204}
\definecolor{steelblue31119180}{RGB}{31,119,180}

\begin{axis}[
legend cell align={left},
legend style={fill opacity=0.8, draw opacity=1, text opacity=1, draw=lightgray204},
tick align=outside,
tick pos=left,
x grid style={darkgray176},
xmin=-0.975, xmax=14.975,
xtick style={color=black},
xtick={0.15,1.15,2.15,3.15,4.15,5.15,6.15,7.15,8.15,9.15,10.15,11.15,12.15,13.15,14.15},
xticklabel style={rotate=90.0},
xticklabels={
  monastery,
  church,
  bell cote,
  palace,
  mosque,
  castle,
  stupa,
  lakeside,
  dome,
  obelisk,
  vault,
  barn,
  book jacket,
  triumphal arch,
  beacon
},
y grid style={darkgray176},
ymin=0, ymax=10719.45,
ytick style={color=black}
]
\draw[draw=none,fill=steelblue31119180] (axis cs:-0.25,0) rectangle (axis cs:-0.05,7251);
\addlegendimage{ybar,ybar legend,draw=none,fill=steelblue31119180}
\addlegendentry{LSUN-Church}

\draw[draw=none,fill=steelblue31119180] (axis cs:0.75,0) rectangle (axis cs:0.95,6731);
\draw[draw=none,fill=steelblue31119180] (axis cs:1.75,0) rectangle (axis cs:1.95,3766);
\draw[draw=none,fill=steelblue31119180] (axis cs:2.75,0) rectangle (axis cs:2.95,3528);
\draw[draw=none,fill=steelblue31119180] (axis cs:3.75,0) rectangle (axis cs:3.95,2047);
\draw[draw=none,fill=steelblue31119180] (axis cs:4.75,0) rectangle (axis cs:4.95,1309);
\draw[draw=none,fill=steelblue31119180] (axis cs:5.75,0) rectangle (axis cs:5.95,679);
\draw[draw=none,fill=steelblue31119180] (axis cs:6.75,0) rectangle (axis cs:6.95,593);
\draw[draw=none,fill=steelblue31119180] (axis cs:7.75,0) rectangle (axis cs:7.95,429);
\draw[draw=none,fill=steelblue31119180] (axis cs:8.75,0) rectangle (axis cs:8.95,371);
\draw[draw=none,fill=steelblue31119180] (axis cs:9.75,0) rectangle (axis cs:9.95,320);
\draw[draw=none,fill=steelblue31119180] (axis cs:10.75,0) rectangle (axis cs:10.95,173);
\draw[draw=none,fill=steelblue31119180] (axis cs:11.75,0) rectangle (axis cs:11.95,163);
\draw[draw=none,fill=steelblue31119180] (axis cs:12.75,0) rectangle (axis cs:12.95,133);
\draw[draw=none,fill=steelblue31119180] (axis cs:13.75,0) rectangle (axis cs:13.95,126);
\draw[draw=none,fill=darkorange25512714] (axis cs:-0.1,0) rectangle (axis cs:0.1,10209);
\addlegendimage{ybar,ybar legend,draw=none,fill=darkorange25512714}
\addlegendentry{DDPM}

\draw[draw=none,fill=darkorange25512714] (axis cs:0.9,0) rectangle (axis cs:1.1,3611);
\draw[draw=none,fill=darkorange25512714] (axis cs:1.9,0) rectangle (axis cs:2.1,3459);
\draw[draw=none,fill=darkorange25512714] (axis cs:2.9,0) rectangle (axis cs:3.1,2103);
\draw[draw=none,fill=darkorange25512714] (axis cs:3.9,0) rectangle (axis cs:4.1,1630);
\draw[draw=none,fill=darkorange25512714] (axis cs:4.9,0) rectangle (axis cs:5.1,1877);
\draw[draw=none,fill=darkorange25512714] (axis cs:5.9,0) rectangle (axis cs:6.1,1406);
\draw[draw=none,fill=darkorange25512714] (axis cs:6.9,0) rectangle (axis cs:7.1,873);
\draw[draw=none,fill=darkorange25512714] (axis cs:7.9,0) rectangle (axis cs:8.1,254);
\draw[draw=none,fill=darkorange25512714] (axis cs:8.9,0) rectangle (axis cs:9.1,488);
\draw[draw=none,fill=darkorange25512714] (axis cs:9.9,0) rectangle (axis cs:10.1,318);
\draw[draw=none,fill=darkorange25512714] (axis cs:10.9,0) rectangle (axis cs:11.1,260);
\draw[draw=none,fill=darkorange25512714] (axis cs:11.9,0) rectangle (axis cs:12.1,154);
\draw[draw=none,fill=darkorange25512714] (axis cs:12.9,0) rectangle (axis cs:13.1,187);
\draw[draw=none,fill=darkorange25512714] (axis cs:13.9,0) rectangle (axis cs:14.1,203);
\draw[draw=none,fill=forestgreen4416044] (axis cs:0.05,0) rectangle (axis cs:0.25,8282);
\addlegendimage{ybar,ybar legend,draw=none,fill=forestgreen4416044}
\addlegendentry{StyleSwin}

\draw[draw=none,fill=forestgreen4416044] (axis cs:1.05,0) rectangle (axis cs:1.25,5711);
\draw[draw=none,fill=forestgreen4416044] (axis cs:2.05,0) rectangle (axis cs:2.25,3204);
\draw[draw=none,fill=forestgreen4416044] (axis cs:3.05,0) rectangle (axis cs:3.25,3331);
\draw[draw=none,fill=forestgreen4416044] (axis cs:4.05,0) rectangle (axis cs:4.25,2569);
\draw[draw=none,fill=forestgreen4416044] (axis cs:5.05,0) rectangle (axis cs:5.25,893);
\draw[draw=none,fill=forestgreen4416044] (axis cs:6.05,0) rectangle (axis cs:6.25,797);
\draw[draw=none,fill=forestgreen4416044] (axis cs:7.05,0) rectangle (axis cs:7.25,672);
\draw[draw=none,fill=forestgreen4416044] (axis cs:8.05,0) rectangle (axis cs:8.25,418);
\draw[draw=none,fill=forestgreen4416044] (axis cs:9.05,0) rectangle (axis cs:9.25,403);
\draw[draw=none,fill=forestgreen4416044] (axis cs:10.05,0) rectangle (axis cs:10.25,445);
\draw[draw=none,fill=forestgreen4416044] (axis cs:11.05,0) rectangle (axis cs:11.25,287);
\draw[draw=none,fill=forestgreen4416044] (axis cs:12.05,0) rectangle (axis cs:12.25,126);
\draw[draw=none,fill=forestgreen4416044] (axis cs:13.05,0) rectangle (axis cs:13.25,105);
\draw[draw=none,fill=forestgreen4416044] (axis cs:14.05,0) rectangle (axis cs:14.25,127);
\end{axis}

\end{tikzpicture}